%% file: main.tex
\documentclass{article}

\usepackage{microtype}
\usepackage{graphicx}
\usepackage{subcaption}
\usepackage{booktabs} 
\usepackage{multirow}
\usepackage{float}
\usepackage{siunitx}
\usepackage{bbm}

\usepackage[accepted]{icml2024}

\usepackage{amsmath}
\usepackage{amssymb}
\usepackage{mathtools}
\usepackage{amsthm}

\usepackage{hyperref}
\usepackage[capitalize,noabbrev]{cleveref}

\theoremstyle{plain}

\theoremstyle{definition}

\theoremstyle{remark}

\newcommand{\model}{\textit{SleepFM}}

\icmltitlerunning{SleepFM: Multi-Modal Sleep Foundation Model}

\begin{document}

\twocolumn[
\icmltitle{SleepFM: Multi-modal Representation Learning for Sleep Across Brain Activity, ECG and Respiratory Signals}



\icmlsetsymbol{equal}{*}



\begin{icmlauthorlist}
\icmlauthor{Rahul Thapa}{dbds}
\icmlauthor{Bryan He}{cs}
\icmlauthor{Magnus Ruud Kjær}{dtu}
\icmlauthor{Hyatt Moore}{psy}
\icmlauthor{Gauri Ganjoo}{psy}
\icmlauthor{Emmanuel Mignot}{psy,senior}
\icmlauthor{James Zou}{dbds,cs,senior}
\end{icmlauthorlist}

\icmlaffiliation{dbds}{Department of Biomedical Data Science, Stanford University}
\icmlaffiliation{cs}{Department of Computer Science, Stanford University}
\icmlaffiliation{dtu}{Department of Health Technology, Technical University of Denmark}
\icmlaffiliation{psy}{Department of Psychiatry and Behavioral Sciences, Stanford University}
\icmlaffiliation{senior}{Co-senior authors}

\icmlcorrespondingauthor{Rahul Thapa}{rthapa84@stanford.edu}

\icmlkeywords{Deep Learning, Foundation Model, Sleep Study}

\vskip 0.3in
]



\printAffiliationsAndNotice{}  

\begin{abstract}

Sleep is a complex physiological process evaluated through various modalities recording electrical brain, cardiac, and respiratory activities. We curate a large polysomnography dataset from over 14,000 participants comprising over 100,000 hours of multi-modal sleep recordings. Leveraging this extensive dataset, we developed \model, the first multi-modal foundation model for sleep analysis. We show that a novel leave-one-out approach for contrastive learning significantly improves downstream task performance compared to representations from standard pairwise contrastive learning. A logistic regression model trained on \model's learned embeddings outperforms an end-to-end trained convolutional neural network (CNN) on sleep stage classification (macro AUROC 0.88 vs 0.72 and macro AUPRC 0.72 vs 0.48) and sleep disordered breathing detection (AUROC 0.85 vs 0.69 and AUPRC 0.77 vs 0.61).  Notably, the learned embeddings achieve 48\% top-1 average accuracy in retrieving the corresponding recording clips of other modalities from 90,000 candidates. This work demonstrates the value of holistic multi-modal sleep modeling to fully capture the richness of sleep recordings. {\model} is open source and available at \href{https://github.com/rthapa84/sleepfm-codebase}{https://github.com/rthapa84/sleepfm-codebase}.

\end{abstract}

%
%

\input{1_Introduction}

%
%

\input{2_Related_Work}

%
%

\input{3_Method}

%
%

\input{4_Experiments_Results}

%


%

\input{5_Discussion_Conclusion}


\section*{Acknowledgements}

RT gratefully acknowledges funding from the Knight-Hennessy Graduate Fellowship. JZ is supported by funding from the Chan-Zuckerberg Biohub. 

\section*{Impact Statement}

Our work develops an initial foundation model for sleep, leveraging multi-modal PSG data from 14,000 sleep studies. We rigorously trained and evaluated against clinically relevant applications to demonstrate impact. To ensure replicability, we release our source code. All of the data used have been deidentified to protect participant Protected Health Information (PHI). 

While our model shows promise, we acknowledge the importance of reliability, transparency, and mitigating potential biases for medical AI. As deep learning is increasingly developed and deployed in sleep medicine, maintaining high ethical standards around consent, explainability, and accountability will be imperative. We believe centering accessibility, responsibility, and social good will allow these technologies to responsibly transform medical practice. Overall, our work is a step toward using AI to capture and unlock the richness of sleep data to better understand and improve our health.

\nocite{langley00}

\bibliography{main}
\bibliographystyle{icml2024}

\newpage
\appendix
\onecolumn

\input{6_Appendix}


\end{document}

%% file: 1_Introduction.tex
\section{Introduction}
\label{section_intro}

Sleep monitoring is critical to evaluate sleep disorders but also as a proxy to assess overall brain, pulmonary, and cardiac health \cite{worley2018extraordinary, brink2022age, leary2021living}. Polysomnography (PSG) is the current gold standard for studying sleep by recording diverse physiological signals during sleep, including electroencephalogram (EEG), electroocculograms (EOG), and electromyography (EMG), electrocardiogram (ECG) and respiratory channels \cite{kryger2010principles}. EOG and EMG are often combined with EEG recordings to better determine sleep stages, which we refer to collectively as Brain Activity Signals (BAS). These different data modalities offer complementary perspectives. BAS measures brain activity to categorize sleep stages and diagnose sleep disorders. ECG captures heart rhythms; changes in heart rate can indicate sleep disordered breathing events. Respiratory sensors directly quantify breathing patterns including sleep disordered breathing (SDB). Together, these signals provide a comprehensive assessment of sleep health.

Traditionally, sleep data analysis involved manual visual inspection, a labor-intensive and time-consuming process prone to errors \cite{boashash2016automatic, hassan2017automated}. Recent advancements in supervised deep learning have shown promise in automating sleep staging and classification of disorders like SDB \cite{nassi2021automated, perslev2021u, stephansen2018neural}. However, most methods rely on labeled data from a narrow task. They rarely leverage the full breadth of unlabeled physiological dynamics within and across diverse PSG sensors.

In parallel, contrastive learning (CL) has emerged as a powerful technique in other domains to learn representations by maximizing alignment between modalities \cite{radford2021learning}. However, joint integration of BAS, ECG, and respiratory signals from PSGs via multi-modal CL has been less explored. Previous works have focused solely on ECG or combined ECG with electronic health records (EHR), while joint modeling of BAS, respiratory, and ECG signals has been limited. Our work demonstrates a first attempt at developing a multi-modal CL approach for PSG analysis that capitalizes on synergies between BAS, ECG, and respiratory signals to learn enhanced physiological representations for sleep analysis.

\textbf{Our Contribution} We introduce {\model}, a sleep foundation model trained using CL on a multi-modal PSG dataset comprising over 100,000 hours of sleep monitoring data from over 14,000 participants at Stanford sleep clinic collected between 1999 and 2020. By combining BAS, ECG, and respiratory modalities from PSG, {\model} exhibits superior performance on tasks such as demographic attributes, sleep stage, and SDB event classifications, outperforming end-to-end trained convolutional neural network (CNN) models. Additionally, we introduce a novel leave-one-out approach for CL, which significantly outperforms the standard pairwise CL on all of our downstream tasks. To our knowledge, this is the first attempt to build and evaluate a multi-modal foundation model for sleep analysis.

%% file: 2_Related_Work.tex
\section{Related Work}
\label{section_related_work}

\subsection{Machine Learning for Analyzing Sleep Data}
\label{section_related_work_sleep_study}

The application of machine learning (ML) in sleep studies has garnered significant recent attention, promising to streamline and expedite the sleep scoring process as well as detecting respiratory events such as SDB. Models including autoencoders \cite{tsinalis2016automatic}, convolutional neural networks (CNNs) \cite{tsinalis1610automatic, sors2018convolutional, yildirim2019deep}, recurrent neural networks (RNNs) \cite{michielli2019cascaded, phan2019seqsleepnet}, and multiple other variations of deep neural networks (DNNs) \cite{supratak2017deepsleepnet, mousavi2019sleepeegnet, seo2020intra, phan2021xsleepnet, perslev2021u} have been proposed for sleep scoring tasks.

Moreover, in the domain of respiratory event classification, automatic detection of SDB using ECG \cite{urtnasan2020automatic, tripathy2020automated}, EEG \cite{zhao2021classification}, and PSG with its respiratory channels \cite{mostafa2020multi, yu2022sleep, haidar2018convolutional, yeo2021respiratory, nassi2021automated, stephansen2018neural} has been explored extensively. A recent study introduced a multi-task learning approach, training a supervised deep learning model to predict diverse sleep events (e.g., sleep stages, arousal, leg movements, and sleep-disordered breathing) using multiple sleep modalities like EEG, EOG, and EMG \cite{zahid2023msed}. These studies predominantly utilize supervised learning, often limited by a narrow subset of downstream tasks.

\subsection{Contrastive Learning}
\label{section_related_work_contrastive_Learning}

A major development in self-supervised learning techniques is the rise of contrastive methods for comprehensive data representation learning. In computer vision, influential frameworks have emerged including: InfoNCE \cite{oord2018representation}, SimCLR \cite{chen2020simple}, MoCo \cite{he2020momentum}, and SupCon \cite{khosla2020supervised}. These uni-modal contrastive approaches focus primarily on single data modalities like images. A notable multi-modal exception is the Contrastive Language-Image Pretraining (CLIP) model \cite{radford2021learning}, which aligns image and text embeddings. In medicine, ConVIRT \cite{zhang2022contrastive} pioneered multi-modal CL between chest radiographs and reports. Other works have explored similar directions for medical images \cite{huang2021gloria, boecking2022making, bannur2023learning, lu2023visual}.

Outside of computer vision, uni-modal contrastive methods have been applied to time series data like ECG signals \cite{kiyasseh2021clocs, gopal20213kg}. CL has also enabled signal conversion tasks \cite{norskov2023cslp}. However, contrastive representation learning across diverse physiological modalities remains relatively uncharted. Two prior studies have investigated contrastive multi-modal clinical time series analysis. One work employed SimCLR-style pre-training on data encompassing ECG and structured records \cite{raghu2022contrastive}. Another derived ECG representations by contrasting ECGs, structured EHRs, and clinical notes \cite{lalam2023ecg}.

{\model} differs from these past works in two primary ways. First, it explores self-supervised representation learning on a large sleep dataset, while most prior works rely on supervised learning. Second, it is the first contrastive model that utilizes a wide array of sleep modalities such as BAS, ECG waveforms, and respiratory signals, covering 19 data channels across three main physiological systems: brain, heart, and lungs. Alongside pairwise CL, we propose and evaluate a novel leave-one-out CL approach. Comprehensive downstream tasks verify {\model}'s superior performance over supervised baseline.

%% file: 3_Method.tex
\section{Method}
\label{section_method}
\subsection{Dataset and Preprocessing}
\label{section_datasets}
Our dataset encompasses PSG records from Stanford Sleep Clinic from 1999-2020, spanning participants aged 2-91. Comprising 14,068 recordings, this dataset features diverse waveforms, such as BAS, ECG, and respiratory channels collected over approximately 8 hours per individual. Its comprehensive nature makes it a valuable and high-quality resource for sleep-related research.

Our preprocessing strategy aimed to make minimal alterations to preserve raw signal characteristics crucial for nuanced pattern recognition. Each recording consists of three modalities: BAS, ECG, and respiratory, encompassing 10, 2, and 7 channels, respectively. The BAS modality includes channels gauging brain activity from various brain regions (frontal, central, occipital), as well as EOG for eye movement and EMG for chin muscle activation. The ECG modality contains channels that measure electrical cardiac function. The respiratory modality includes channels measuring chest and abdomen movements, pulse readings, nasal and oral flow measurements. The selection of these channels was guided by sleep experts due to their relevance in sleep studies, facilitating sleep stage scoring and SDB detection \cite{berry2012aasm}.

Subsequently, we segmented the total sleep duration into consecutive 30-second clips for all participants, following the standard clip duration used in sleep studies \cite{berry2012aasm}. We then resampled the dataset to 256 Hz to standardize the sampling rate across all participants. Furthermore, expert sleep technicians labeled each clip for both sleep stage and SDB. Sleep stage is categorized into Wake, Stage 1, Stage 2, Stage 3, REM, and SDB is a binary label. To prevent data leakage, the dataset is split into participant-level pretrain/train/validation/test sets consisting of 11,261, 1,265, 141, and 1,401 participants respectively. Each participant contributes multiple clips to our dataset, resulting in a total of 10.6M, 1.19M, 130K, and 1.31M clips, respectively. The pretrain dataset is only used to pretrain our foundation model. The remaining set serves to train and test our model and baseline models for downstream applications as explained in \Cref{section_experiments}. The validation set is used to optimize the hyperparameters. Demographic statistics for different splits are presented in \Cref{tab:demographics}. An illustrative snapshot of our data can be found in \Cref{fig:raw_data}.

\begin{table*}[htbp]
    \centering
    \caption{Demographics table. REM: Rapid Eye Movement; AHI: Apnea-Hypopnea Index, a measure used in sleep medicine to assess the severity of sleep apnea; WASO: Wake After Sleep Onset, the total time spent awake after initially falling asleep; SL: Sleep Latency, the time it takes to transition from wakefulness to sleep; REML: REM Sleep Latency, the time it takes to enter REM sleep after falling asleep; TSD: Total Sleep Duration, the overall duration of sleep. $\pm$ represents upper and lower bound.}
    \label{tab:measurement_results}
    
    \begin{tabular}{@{}lcccc@{}}
        \toprule
        & pretrain & train & valid & test \\
        \midrule
        Participants (count) & 11,261 & 1,265 & 141 & 1,401 \\
        Events (count) & 10,611,314 & 1,190,392 & 130,380 & 1,314,267 \\
        Duration (hours) & 88,427 & 9,920 & 1,086 & 10,952 \\
        \midrule
        Male (\%) & 49.9 & 50.2 & 47.1 & 53.0 \\
        Female (\%) & 43.8 & 44.0 & 48.1 & 41.8 \\
        Unknown (\%) & 6.3 & 5.9 & 4.8 & 5.2 \\
        Age (years) & 42.2 $\pm$ 19.6 & 43.0 $\pm$ 20.3 & 40.4 $\pm$ 20.0 & 41.9 $\pm$ 19.9 \\
        \midrule
        TSD (mins) & 376.7 $\pm$ 90.8 & 376.4 $\pm$ 90.6 & 371.2 $\pm$ 84.9 & 374.3 $\pm$ 87.5 \\
        WASO (mins) & 79.4 $\pm$ 60.5 & 79.7 $\pm$ 62.3 & 78.8 $\pm$ 57.3 & 81.5 $\pm$ 62.8 \\
        SL (mins) & 22.2 $\pm$ 32.8 & 21.2 $\pm$ 31.6 & 29.0 $\pm$ 87.8 & 22.5 $\pm$ 32.6 \\
        REML (mins) & 151.9 $\pm$ 102.6 & 149.4 $\pm$ 97.7 & 148.6 $\pm$ 99.9 & 154.8 $\pm$ 103.5 \\
        \midrule
        Stage 1 (\%) & 9.4 $\pm$ 9.2 & 9.3 $\pm$ 8.8 & 8.2 $\pm$ 7.7 & 9.0 $\pm$ 8.9 \\
        Stage 2 (\%) & 65.0 $\pm$ 14.7 & 64.8 $\pm$ 14.7 & 64.8 $\pm$ 14.7 & 65.0 $\pm$ 14.7 \\
        Stage 3 (\%) & 10.2 $\pm$ 13.2 & 10.2 $\pm$ 13.2 & 10.9 $\pm$ 12.7 & 10.3 $\pm$ 13.6 \\
        REM (\%) & 15.5 $\pm$ 7.9 & 15.7 $\pm$ 8.0 & 16.2 $\pm$ 6.8 & 15.7 $\pm$ 7.9 \\
        \midrule
        AHI (h$^{-1}$) & 22.2 $\pm$ 79.3 & 22.8 $\pm$ 19.1 & 22.2 $\pm$ 18.5 & 20.9 $\pm$ 17.0 \\
        \bottomrule
    \end{tabular}
    \label{tab:demographics}
\end{table*}

\subsection{Embedding Model}
\label{section_embedding_model}

Our pre-training stage employed CL as the foundational algorithm for representation learning, explained in more detail in \cref{section_contrastive_learning}. We used three 1D CNNs to generate three separate embeddings from the BAS, ECG, and respiratory modalities and trained them separately. The architecture of the models is based on a 1D CNN developed for classifying ECG measurements \cite{ouyang2022electrocardiographic}.
These models differ in their first convolutional layers to accommodate the number of channels specific to each modality: 10 for BAS, 2 for ECG, and 7 for respiratory channels.

The architecture of these embedding models is rooted in EfficientNet architecture \cite{tan2019efficientnet}. The architecture starts with atrous convolutions followed by subsequent multi-channel 1D convolutions. The layer count aligns with the original design of EfficientNet \cite{tan2019efficientnet}, but the number of channels is significantly reduced for model runtime efficiency and to minimize complexity. Following the initial atrous layers, the model incorporates convolutional layers utilizing an invested residual structure, mirroring the input and output bottleneck layers with an intermediate expansion layer \cite{sandler2018mobilenetv2}.

For regularization, a dropout layer precedes the final fully connected output layer. Depthwise separable convolutions are extensively utilized to minimize parameters while preserving representational capacity. Residual connections aid gradient flow across multiple layers during optimization, facilitating hierarchical feature learning on variable-length sequential data.

\subsection{Multi-modal Contrastive Learning}
\label{section_contrastive_learning}

\begin{figure}[htbp]
    \centering
    \includegraphics[width=1\linewidth]{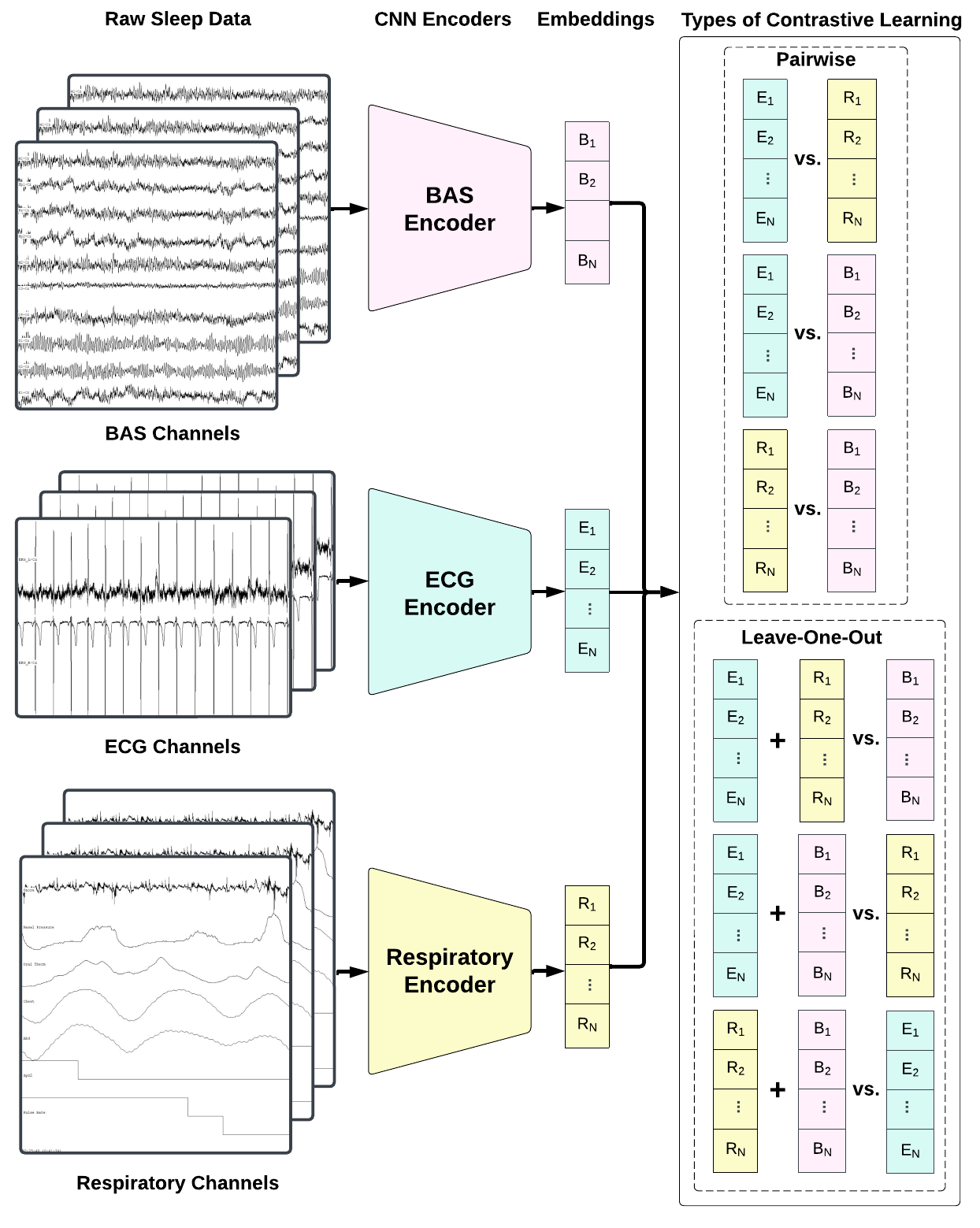}
    \caption{Overview of {\model} pre-training with CL. We experiment with two types of pre-training: standard pairwise CL where we contrast embeddings from each pair of modalities separately, and our novel leave-one-out CL where we contrast the embedding of each modality against the average embedding of all other modalities. BAS (Brain Activity Signals) measures brain activity, eye and muscle movement, Electrocardiogram (ECG) measures heart activity, and Respiratory channels measure chest, abdomen movements, pusle, nasal, and oral flow.}
    \label{fig:architecture_fig}
\end{figure}

We explore two CL frameworks for learning joint representations across modalities: pairwise CL and leave-one-out CL (  \cref{fig:architecture_fig}). The key idea is to bring positive pairs of embeddings from different modalities closer in the latent space while pushing apart negative pairs. The positive pairs are derived from temporally aligned 30-second clips across modalities. All other non-matching instances within a training batch are treated as negative pairs.

In pairwise CL, we construct contrastive prediction tasks between all pairs of modalities.
We use a contrastive loss to encourage agreement between positive pairs while discouraging agreement between negative pairs.
Specifically, for modalities $i$ and $j$ and sample $k$ in a batch, we have an embedding $x_k^i$ from modality $i$ and an embedding $x_k^j$ from modality $j$.
The contrastive prediction loss is defined as:
\begin{equation}
l_{i, j, k}^{\mbox{pair}} = -\log \frac{\exp(\mathrm{sim}(x_{k}^{i}, x_{k}^{j}) * \exp(\tau))}{\sum_{m=1}^{N} \exp(\mathrm{sim}(x_{k}^{i}, x_{m}^{j}) * \exp(\tau))},
\end{equation}

where $N$ is the number of samples in a batch, $\tau$ is a trainable temperature parameter, and $\mathrm{sim}$ is cosine similarity. We sum this loss over all the samples in a batch and repeat the process for all pairs of modalities ${i, j}$. The final loss is the sum of pairwise contrastive losses over all modality pairs.

In leave-one-out CL, we construct a predictive task where an embedding from one modality tries to identify the corresponding embeddings from the remaining modalities. In particular, for each modality $i$, we construct an embedding $\bar{x}^{\neq i}$ by averaging over embeddings from all other modalities, excluding modality $i$. We then apply a contrastive loss between modality $i$'s embedding and this leave-one-out representation:
\begin{equation}
l_{i, k}^{\mbox{LOO}} = -\log \frac{\exp(\text{sim}(x_{k}^{i}, \overline{x}_{k}^{\neq i})* \exp(\tau))}{\sum_{m=1}^{N} \exp(\text{sim}(x_{k}^{i}, \overline{x}_{m}^{\neq i})* \exp(\tau))}
\end{equation}

Similar to pairwise, this is the loss for a sample $k$ from modality ${i}$ in a given batch. 

The motivation behind the leave-one-out method is to encourage each embedding to capture semantics aligned with all other modalities. Pairwise CL, on the other hand, encourages alignments only between particular pairs of modalities.

\subsection{Model Training}
\label{subsection_model_training}

Our model pretraining involves minimizing the contrastive loss with stochastic gradient descent (SGD) using an initial learning rate set to 0.001 and a momentum of 0.9. The learning rate is decayed by a factor of 10 every 5 epochs. The trainable temperature parameter is initialized to 0. Training spans a maximum of 20 epochs with early stopping based on validation loss, employing a batch size of 32 and validating checkpoints at each epoch to ensure robust regularization.

Upon pretraining completion via this self-supervised approach, we generate embeddings for the training, validation, and test sets, utilizing the learned modality encoders. Subsequently, these training embeddings drive the training of a logistic regression classifier. The classifier's performance undergoes evaluation on the test set for both sleep stage and SDB event detection tasks, as outlined in \cref{subsection_downstream_tasks}.

In our experiments, we additionally compare against training a supervised CNN without contrastive learning as a baseline.
The supervised CNN uses an 1D EfficientNet architecture akin to our pretrained model encoder but is solely trained via supervised learning on the entire (pretraining + training) dataset for classification tasks. This architecture uses a series of 1D convolutions encoding all three modalities into an embedding space, followed by a dropout layer for regularization and a fully-connected layer predicting scores across different classes. This model is trained end-to-end from scratch using cross-entropy loss between the predicted and true labels, optimized by SGD. Mirroring the pretraining phase, this model undergoes training for 20 epochs with a batch size of 32, aligning hyperparameters with our model pretraining strategy. Additional training details are available in \cref{subsection_training_details_supplementary}.

%% file: 4_Experiments_Results.tex
\section{Experiments and Results}
\label{section_experiments}

\subsection{Demographic Attributes Classification}
\label{subsection_demographics_classification} 

\begin{table*}[t]
    \centering
    \caption{Age classification metrics for models trained using different types of contrastive learning (CL). The supervised CNN is trained on the entire (pretraining + training) dataset to classify age groups. The leave-one-out and pairwise models are logistic regression models trained on the embeddings generated from only the training dataset. Therefore end-to-end CNN saw all data 11,261 participants while pretrained model saw data from 1,265 participants for sleep stage classification. Prevalence of 0-18, 18-35, 35-50, and 50+ are 0.17, 0.18, 0.28, and 0.37 respectively. $\pm$ represents 95\% Confidence Intervals.}
    \label{tab:age_classification_metrics_contrastive_learning}
    \begin{tabular}{@{}lccc|ccc@{}}
        \toprule
        & \multicolumn{3}{c|}{\textbf{AUROC}} & \multicolumn{3}{c}{\textbf{AUPRC}} \\
        \cmidrule(r){2-4} \cmidrule(l){5-7}
        & \textbf{Leave-One-Out} & \textbf{Pairwise} & \textbf{Supervised CNN} & \textbf{Leave-One-Out} & \textbf{Pairwise} & \textbf{Supervised CNN}\\
        \midrule
        0-18 & $0.982_{\pm .001}$  & $0.977_{\pm .001}$ & $0.864_{\pm .001}$ & $0.937_{\pm .002}$ & $0.929_{\pm .004}$ & $0.628_{\pm .003}$\\
        18-35 & $0.852_{\pm .001}$ & $0.809_{\pm .002}$ & $0.683_{\pm .002}$ & $0.549_{\pm .003}$ & $0.458_{\pm .002}$ & $0.308_{\pm .002}$\\
        35-50 & $0.784_{\pm .001}$  & $0.740_{\pm .001}$ & $0.606_{\pm .003}$ & $0.524_{\pm .001}$ & $0.476_{\pm .002}$ & $0.371_{\pm .002}$\\
        50+ & $0.915_{\pm .001}$  & $0.880_{\pm .001}$ & $0.745_{\pm .002}$ & $0.856_{\pm .002}$ & $0.796_{\pm .002}$ & $0.619_{\pm .002}$\\
        \midrule
        \textbf{Avg} & $\mathbf{0.883}$ & $0.851$ & $0.724$ & $\mathbf{0.716}$ & $0.664$ & $0.481$\\
        \bottomrule
    \end{tabular}
\end{table*}

\begin{table}[t]
    \centering
    \caption{Gender classification metrics for models trained using different types of CL. The supervised CNN is trained on the entire (pretraining + training) dataset to classify gender. The leave-one-out and pairwise models are logistic regression models trained on the embeddings generated from only the training dataset. Therefore end-to-end CNN saw 11,261 patient data while pretrained model saw 1,265 training data for SDB classification. Prevalence of female gender is 0.41. $\pm$~represents 95\% Confidence Intervals.}
    \label{tab:gender_classification_metrics_contrastive_learning}
    \begin{tabular}{@{}lcc@{}}
        \toprule
        & \textbf{AUROC} & \textbf{AUPRC}\\
        \midrule
        \textbf{Leave-One-Out CL} & $\mathbf{0.850_{\pm .001}}$ & $\mathbf{0.774_{\pm .002}}$ \\
        \textbf{Pairwise CL} & $0.810_{\pm .001}$ & $0.731_{\pm .002}$ \\
        \textbf{Supervised CNN} & $0.690_{\pm .002}$ & $0.614_{\pm .002}$ \\
        \bottomrule
    \end{tabular}
\end{table}

We evaluated our {\model}'s embedding quality by training a logistic regression classifier on top of the combined multimodal embeddings to predict common demographic attributes such as age and gender. Our classification task directly used the 30-second clip-level embeddings generated by {\model}. For age prediction, we grouped ages into the following categories: 0-18, 18-35, 35-50, and 50+. The prevalence of these age groups in our dataset is 0.17, 0.18, 0.28, and 0.37, respectively. For gender classification, we considered male vs. female, with the prevalence of females being 0.41 in our dataset. We evaluated the performance based on AUROC (Area Under the Receiver Operating Characteristic curve) and AUPRC (Area Under the Precision-Recall Curve).
As a baseline, we trained a CNN end-to-end to perform age and gender classification given the combined multimodal raw input data.

We find that {\model} can predict age and gender with high accuracy from just 30-second clips of physiological data
(\cref{tab:age_classification_metrics_contrastive_learning} and \cref{tab:gender_classification_metrics_contrastive_learning}).
Both our pre-trained models significantly outperform the end-to-end CNN baseline across all evaluation metrics and tasks.
Note that the end-to-end supervised CNN used the full (pretraining + training) dataset during training, while the embeddings from {\model} were only trained on the training set.
Notably, the model pre-trained with leave-one-out CL achieves the best performance.
The strong clip-level performance indicates {\model}'s embeddings effectively capture salient demographic information.
Analyzing the performance per modality, we find that the BAS signals contain the most distinctive features for these tasks as shown in \cref{tab:modality_age_classification_metrics_leave_one_out} and \cref{tab:modality_gender_classification_metrics_leave_one_out}.

\subsection{Retrieval Analysis}
\label{subsection_retrieval_analysis} 

\begin{table}[t]
\centering
\caption{Retrieval on the test set for model trained with leave-one-out contrastive learning (CL). Resp is for Respiratory. Random baseline for Recall@10 = 0.0001}
\begin{tabular}{@{}lcccccc@{}}
  \toprule
  & \multicolumn{3}{c}{\textbf{Median Rank}} & \multicolumn{3}{c}{\textbf{Recall@10}} \\
  \cmidrule(lr){2-4}\cmidrule(lr){5-7}
  & BAS & ECG & Resp & BAS & ECG & Resp \\
  \midrule  
  BAS & - & 7 & 416 & - & 0.58 & 0.05 \\
  ECG & 13 & - & 19 & 0.46 & - & 0.39 \\
  Resp & 400 & 21 & - & 0.05 & 0.38 & - \\
  \bottomrule
\end{tabular}
\label{tab:leave_one_out_retrieval_metrics}


\centering
\caption{Retrieval on the test set for model trained with pairwise contrastive learning (CL). Resp is for Respiratory. Random baseline for Recall@10 = 0.0001}
\begin{tabular}{@{}lcccccc@{}}
  \toprule
  & \multicolumn{3}{c}{\textbf{Median Rank}} & \multicolumn{3}{c}{\textbf{Recall@10}} \\
  \cmidrule(lr){2-4}\cmidrule(lr){5-7}
  & BAS & ECG & Resp & BAS & ECG & Resp \\
  \midrule
  BAS & - & 1 & 6 & - & 0.74 & 0.58 \\
  ECG & 1 & - & 2 & 0.82 & - & 0.81 \\
  Resp & 5 & 2 & - & 0.60 & 0.82 & - \\
  \bottomrule   
\end{tabular}
\label{tab:pairwise_retrieval_metrics}
\end{table}

To further assess the quality of {\model}'s embeddings, we assessed its retrieval capabilities by retrieving one modality's closest embeddings from the test set based on another modality's embeddings. For instance, computing cosine similarity between BAS and ECG embeddings generated a ranked list, allowing us to gauge retrieval performance. Evaluation was measured using recall@10 and median rank metrics. 

\begin{itemize}
    \item \textbf{Recall@10}: Measures the true paired item's appearance within the top 10 recommendations. Higher values indicate more accurate retrieval among top recommendations.
    \item \textbf{Median rank}: Determines the median position of the true paired item in rankings; a lower median rank signifies a more consistent ranking of the correct pair among recommendations.
\end{itemize}

We measured the retrieval performance using 90,000 randomly selected 30-second clips encompassing all modalities from the test set. To ensure a representative sample, we uniformly selected clips from various event types across all participants within the test set. The Recall@10 for random retrievals is $10/90000 = 0.0001$.

{\model} achieved over 500x-8000x higher Recall@10 than the random chance as shown in \cref{tab:leave_one_out_retrieval_metrics} and \cref{tab:pairwise_retrieval_metrics}. Pairwise CL yields better overall retrieval performance than leave-one-out, likely because the retrieval evaluation directly maps the training procedure of pairwise. One trend across both metrics is that retrieval performance between respiratory and other modalities is comparatively worse. The discrepancy in retrieval performance may stem from the higher variablilty of the respiratory measurements. While BAS is directly measured via electrical activity from the brain and ECG is directly measured via electrical activity from the heart, the respiratory channels indirectly measure breathing through the movement of the participant, which can be influenced by body position and non-breathing related motion.

\subsection{Downstream Classification Tasks}
\label{subsection_downstream_tasks} 

\begin{table*}[t]
    \centering
    \caption{Sleep stage classification metrics for models trained using different types of contrastive learning (CL). Baseline here is an end-to-end CNN trained on the entire (pretraining + training) dataset to classify sleep stages. The leave-one-out and pairwise models are logistic regression models trained on the embeddings generated from only the training dataset. Therefore end-to-end CNN saw 11,261 patient data while pretrained model saw 1,265 training data for sleep stage classification. Prevalence of Wake, Stage 1, Stage 2, Stage 3, and REM are 0.21, 0.07, 0.51, 0.09, and 0.12 respectively. $\pm$ represents 95\% Confidence Intervals.}
    \label{tab:sleep_stage_classification_metrics_contrastive_learning}
    \begin{tabular}{@{}lccc|ccc@{}}
        \toprule
        & \multicolumn{3}{c|}{\textbf{AUROC}} & \multicolumn{3}{c}{\textbf{AUPRC}} \\
        \cmidrule(r){2-4} \cmidrule(l){5-7}
        & \textbf{Leave-One-Out} & \textbf{Pairwise} & \textbf{Supervised CNN} & \textbf{Leave-One-Out} & \textbf{Pairwise} & \textbf{Supervised CNN}\\
        \midrule
        Wake & $0.945_{\pm .001}$  & $0.930_{\pm .001}$ & $0.869_{\pm .001}$ & $0.862_{\pm .002}$ & $0.827_{\pm .002}$ & $0.711_{\pm .002}$\\
        Stage 1 & $0.814_{\pm .002}$ & $0.782_{\pm .002}$ & $0.706_{\pm .002}$ & $0.233_{\pm .003}$ & $0.186_{\pm .002}$ & $0.130_{\pm .002}$\\
        Stage 2 & $0.891_{\pm .001}$  & $0.861_{\pm .001}$ & $0.840_{\pm .001}$ & $0.876_{\pm .001}$ & $0.849_{\pm .001}$ & $0.822_{\pm .001}$\\
        Stage 3 & $0.928_{\pm .001}$ & $0.918_{\pm .001}$ & $0.918_{\pm .001}$ & $0.676_{\pm .003}$ & $0.615_{\pm .003}$ & $0.695_{\pm .002}$\\
        REM & $0.951_{\pm .001}$  & $0.891_{\pm .001}$ & $0.878_{\pm .001}$ & $0.778_{\pm .003}$ & $0.565_{\pm .002}$ & $0.540_{\pm .003}$\\
        \midrule
        \textbf{Avg} & $\mathbf{0.906}$ & $0.876$ & $0.842$ & $\mathbf{0.685}$ & $0.608$ & $0.579$\\
        \bottomrule
    \end{tabular}
\end{table*}

\begin{table}[t]
    \centering
    \caption{SDB classification metrics for models trained using different types of contrastive learning (CL). Baseline here is a supervised CNN trained on the entire (pretraining + training) dataset to classify SDB. The leave-one-out and pairwise models are logistic regression models trained on the embeddings generated from only the training dataset. Therefore end-to-end CNN saw 11,261 patient data while pretrained model saw 1,265 training data for SDB classification. Prevalence of SDB event is 0.017. $\pm$ represents 95\% Confidence Intervals.}
    \label{tab:SDB_event_classification_metrics_contrastive_learning}
    \begin{tabular}{@{}lcc@{}}
        \toprule
        & \textbf{AUROC} & \textbf{AUPRC}\\
        \midrule
        \textbf{Leave-One-Out CL} & $\mathbf{0.941_{\pm .002}}$ & $\mathbf{0.711_{\pm .006}}$ \\
        \textbf{Pairwise CL} & $0.902_{\pm .003}$ & $0.586_{\pm .007}$ \\
        \textbf{Supervised CNN} & $0.843_{\pm .002}$ & $0.555_{\pm .005}$ \\
        \bottomrule
    \end{tabular}
\end{table}

Having demonstrated that {\model} learns useful representations from PSG clips for tasks such as demographic prediction and clip retrieval, we now evaluate performance on clinically useful downstream tasks: sleep stage and SDB classification. Manual sleep stage scoring and SDB classification currently requires extensive analysis by trained technicians, motivating automatic techniques. To do so, we used the embeddings learned by {\model} to train a logistic regression model and classify sleep stages and SDB events on a held-out test dataset. Sleep stage classification is a multi-class classification task, with 5 classes: Wake, Stage 1, Stage 2, Stage 3, and REM. Prevalence of these groups are 0.21, 0.07, 0.51, 0.09, and 0.12 respectively. SDB classification is a binary classification task, with a prevalence of 0.017. We compared {\model} performance with end-to-end CNN trained on all three modalities, for sleep stage and SDB event classification. 

The results for sleep stage classification are presented in \cref{tab:sleep_stage_classification_metrics_contrastive_learning}. Notably, across both AUROC and AUPRC metrics, the logistic regression model trained using representations from {\model} outperforms the CNN trained end-to-end in a supervised manner. This superiority holds true across all sleep stage classes as well as on aggregated class metrics. Model pretrained with leave-one-out CL performs better than the one pretrained with pairwise across both metrics.

Similarly, the SDB classification metrics, displayed in \cref{tab:SDB_event_classification_metrics_contrastive_learning}, underscore our approach's superiority over supervised CNN models. We find that the model pretrained with leave-one-out CL significantly outperforms the model pretrained with pairwise. While our classification performance aligns with existing methods \cite{salari2022detection, li2022deep}, our study emphasizes the potential of multi-modal CL in these specific domains.

Furthermore, we sought to understand the performance of individual modality embeddings when trained separately for these tasks. \cref{tab:modality_sleep_stage_classification_metrics_leave_one_out} and \cref{tab:modality_SDB_classification_metrics_leave_one_out}, exhibit the results for sleep staging and SDB classification using each modality's embeddings independently. As expected, model trained on BAS embeddings excel in sleep stage classification, while the model trained on respiratory embeddings perform notably well in SDB event detection, as these are the modalities commonly used for the respective tasks. Surprisingly, across both tasks, embeddings from all modalities demonstrated reasonably high performance, specially for sleep stage classification.

Additionally, we stratified the performance of our model across different age and gender groups to ensure there were no discrepancies across demographics. In \cref{tab:sleep_stage_classification_age_macro} and \cref{tab:sleep_stage_classification_gender_macro}, we see that both our pretrained models perform consistently well across all age groups with minor variation, especially among the 50+ age group. Across genders, the performance was similarly consistent with even less variation. For SDB classification, the performance was consistently strong across age and gender groups, except for the 0-18 age group, which exhibited slightly lower performance than other groups as shown in Tables \ref{tab:SDB_classification_age} and \ref{tab:SDB_classification_gender}.

\begin{figure*}[t]
    \centering
    \begin{subfigure}{0.246\textwidth}
        \centering
        \includegraphics[width=\linewidth]{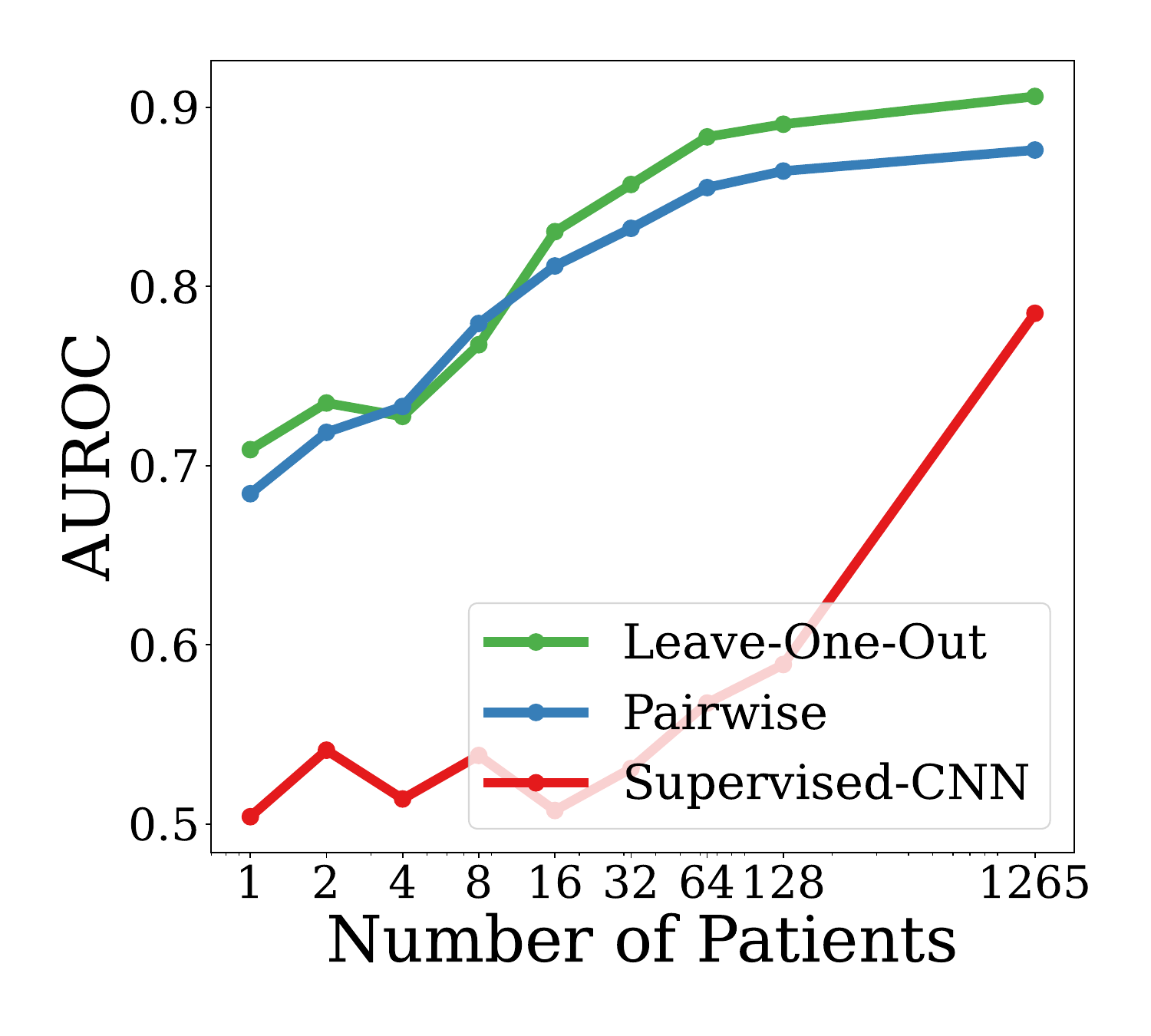}
        \caption{Sleep Stages AUROC}
    \end{subfigure}
    \begin{subfigure}{0.246\textwidth}
        \centering
        \includegraphics[width=\linewidth]{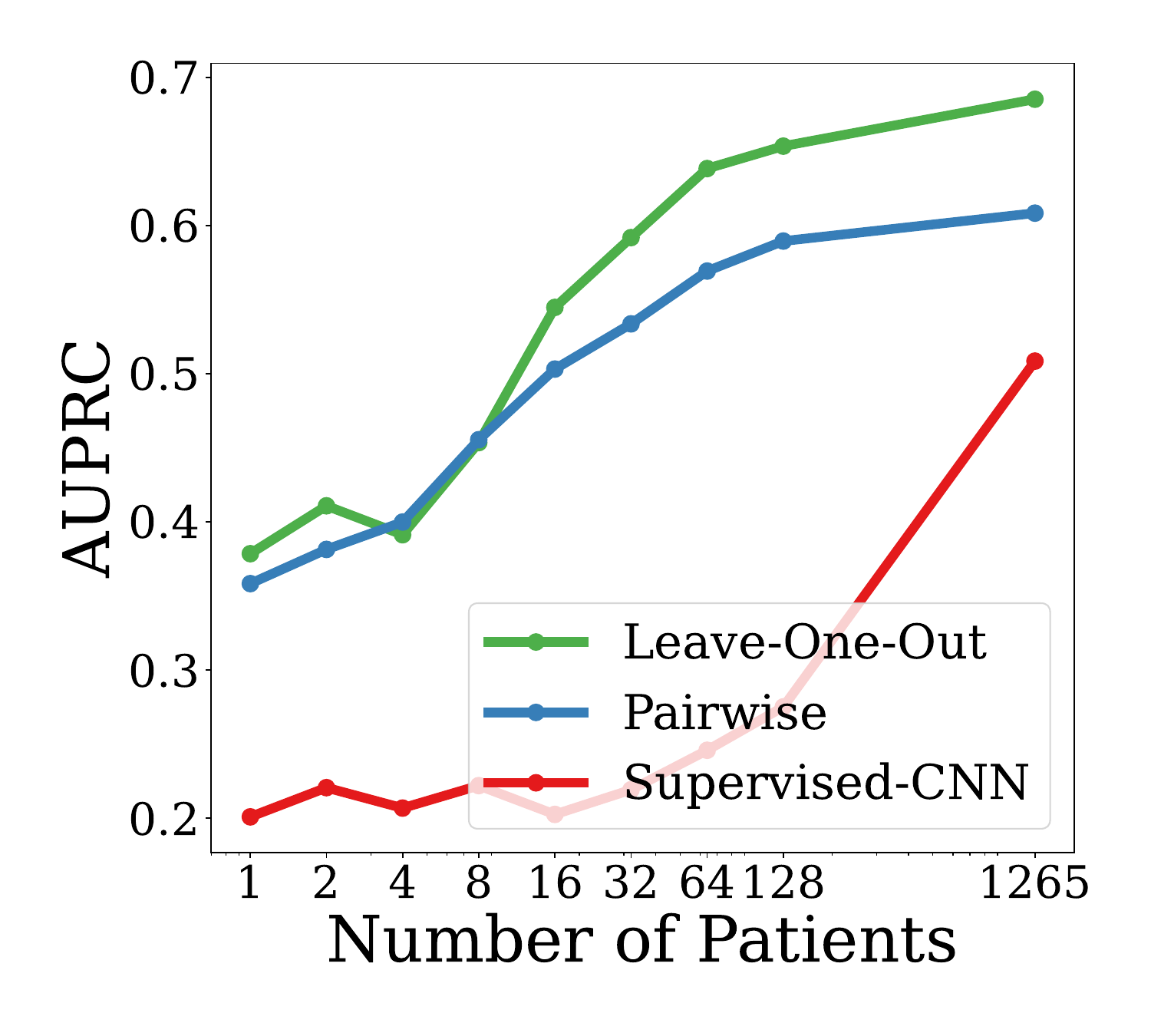}
        \caption{Sleep Stages AUPRC}
    \end{subfigure}
    \begin{subfigure}{0.246\textwidth}
        \centering
        \includegraphics[width=\linewidth]{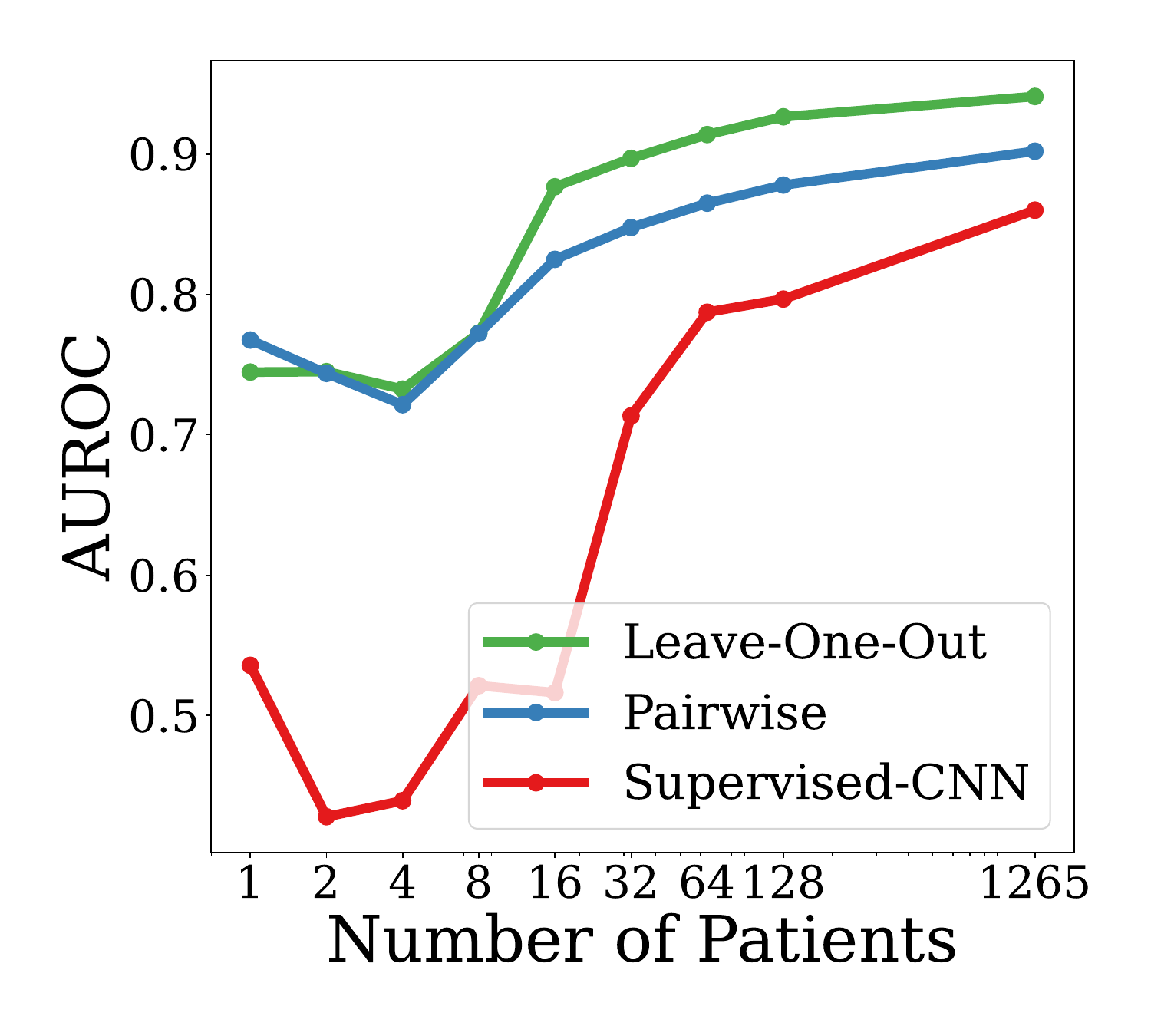}
        \caption{SDB AUROC}
    \end{subfigure}
    \begin{subfigure}{0.246\textwidth}
        \centering
        \includegraphics[width=\linewidth]{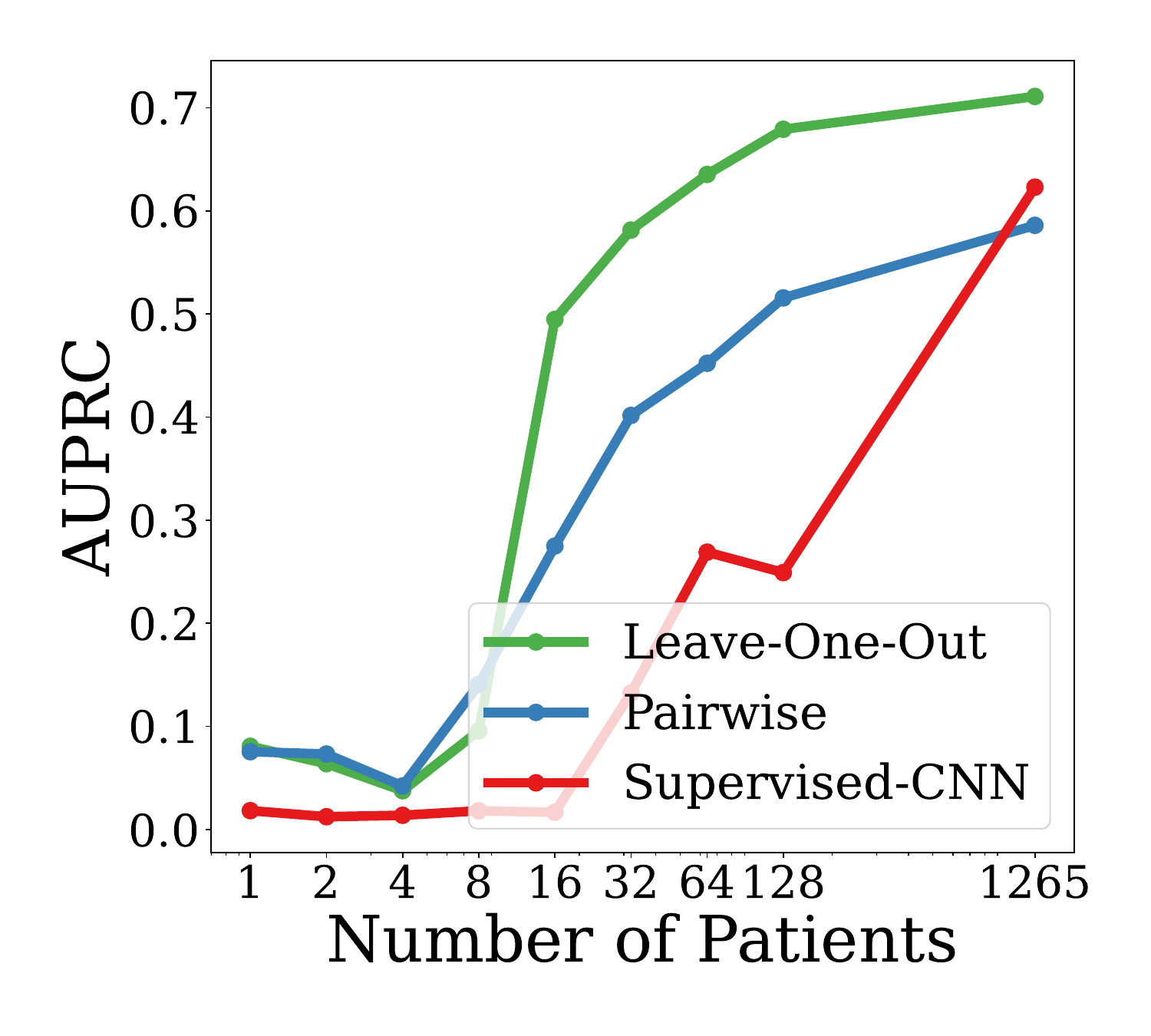}
        \caption{SDB AUPRC}
    \end{subfigure}
    \caption{Few Shot Evaluation. The x-axis represents number of patients that the model was trained on and y-axis represents evaluation metrics AUROC and AUPRC. In case of pairwise and leave-one-out, we select embeddings from $k$ number of patients to train a logistic regression model. The largest number of patients used (1265) is the total size of our training dataset. In case of supervised CNN, we train the model end-to-end on $k$ number of patients to classify either sleep stages or SDB. Testing is done on the entire test set. For each shot, we average the performance across 3 replicates.}
    \label{fig:few_shot_plots}
\end{figure*}

\subsection{Few-Shot Evaluation}
\label{subsection_few_shot_evaluation}

To understand how our model performs when we only have a small sample size available to train a model for downstream application, we performed a few-shot performance evaluation. To do so, we steadily increased the number of participants $k$ that each model sees from $k = 1$ to the full training dataset, and recorded the model's AUROC and AUPRC at each $k$. Note that each participant contributes multiple training clips. We consider values of $k \in \{ 1, 2, 4, 8, 16, 32, 64, 128, 1265\}$, where 1265 is the size of the full training set. For the supervised CNN, few-shot examples are the only training examples seen by the model. For the pretrained models, we use embeddings of these few-shot examples to train a logistic regression model. 

For both AUROC and AUPRC, we see that across all training set sizes, {\model} significantly outperforms baseline supervised CNN model for both sleep stage and SDB classification (\cref{fig:few_shot_plots}). Notably, the leave-one-out model significantly outperforms pairwise model across all training set sizes, especially for SDB classification.

\subsection{Benefit of Multi-Modal Pretraining}
\label{subsection_ablation_modalities} 

\begin{figure*}[t]
    \centering
    \begin{subfigure}{0.246\textwidth}
        \centering
        \includegraphics[width=\linewidth]{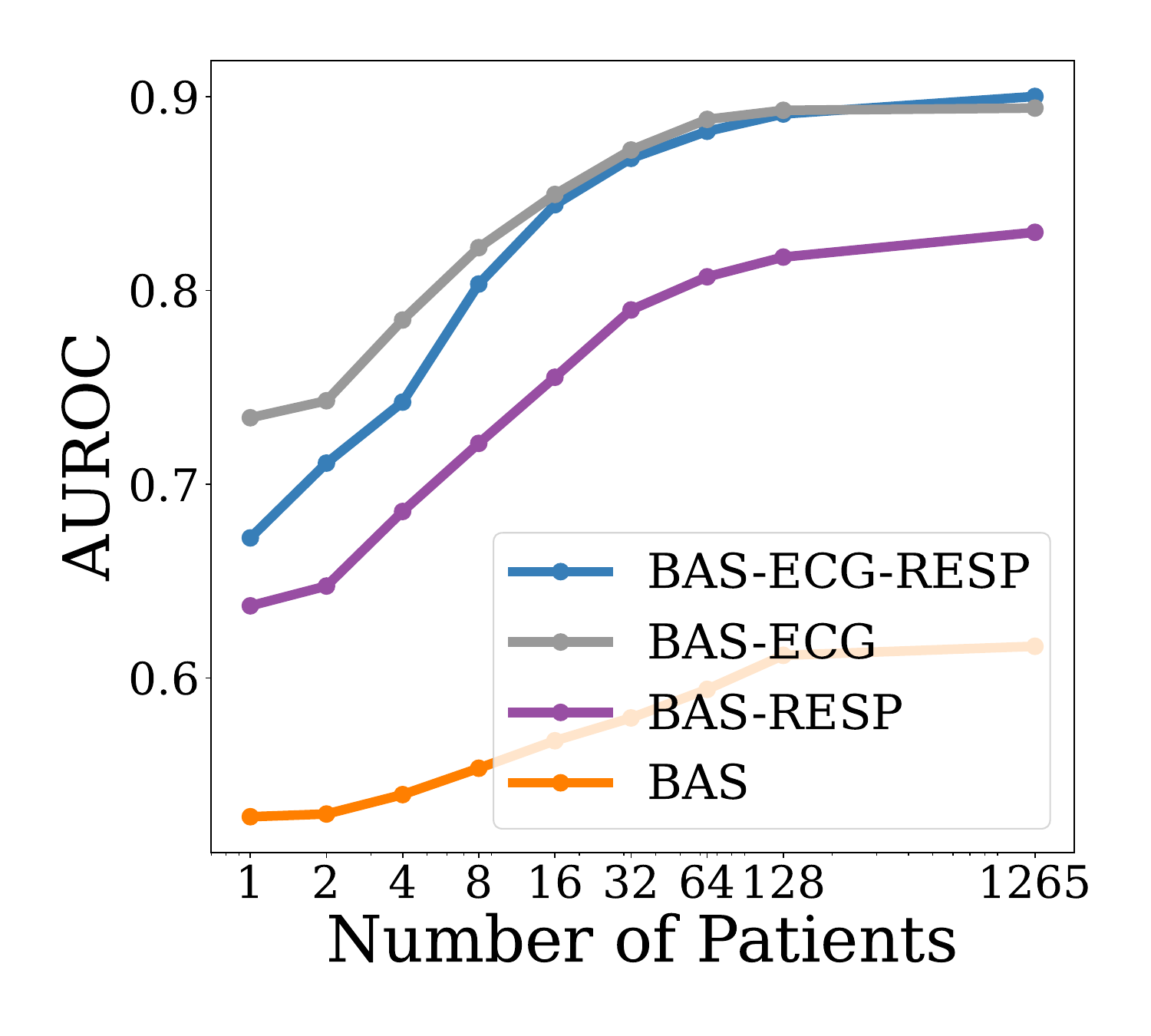}
        \caption{Sleep Stages AUROC}
    \end{subfigure}
    \begin{subfigure}{0.246\textwidth}
        \centering
        \includegraphics[width=\linewidth]{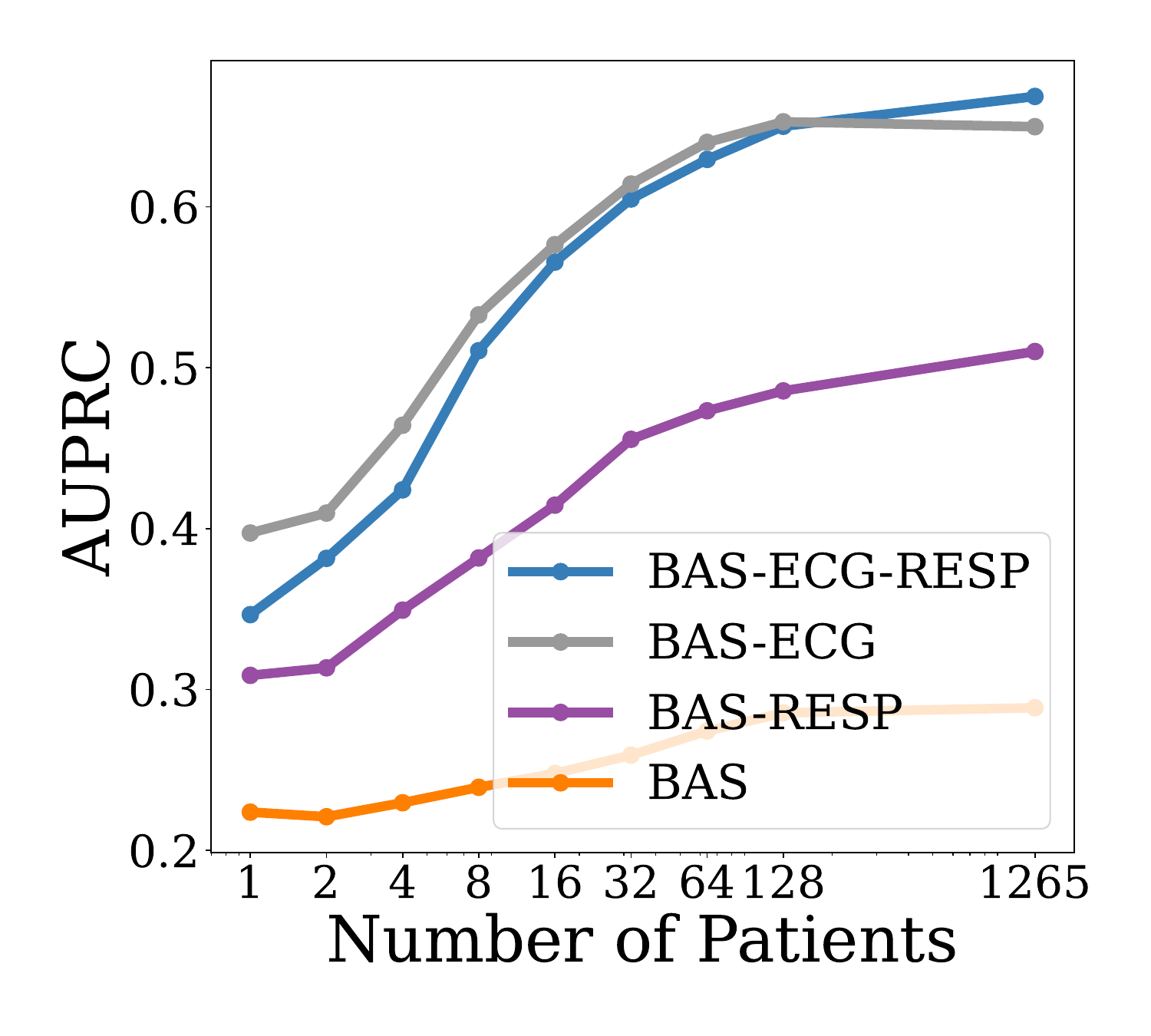}
        \caption{Sleep Stages AUPRC}
    \end{subfigure}
    \begin{subfigure}{0.246\textwidth}
        \centering
        \includegraphics[width=\linewidth]{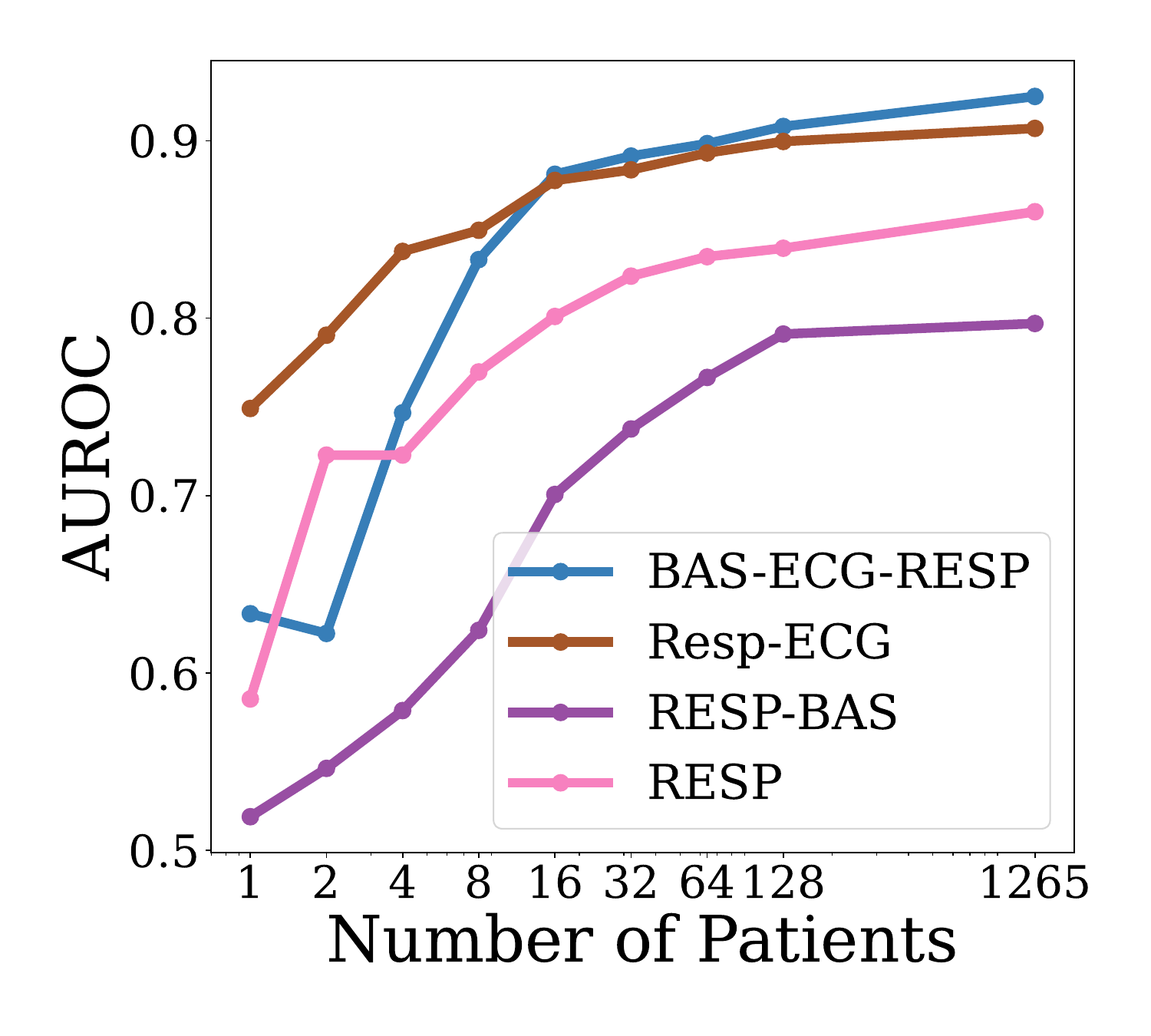}
        \caption{SDB AUROC}
    \end{subfigure}
    \begin{subfigure}{0.246\textwidth}
        \centering
        \includegraphics[width=\linewidth]{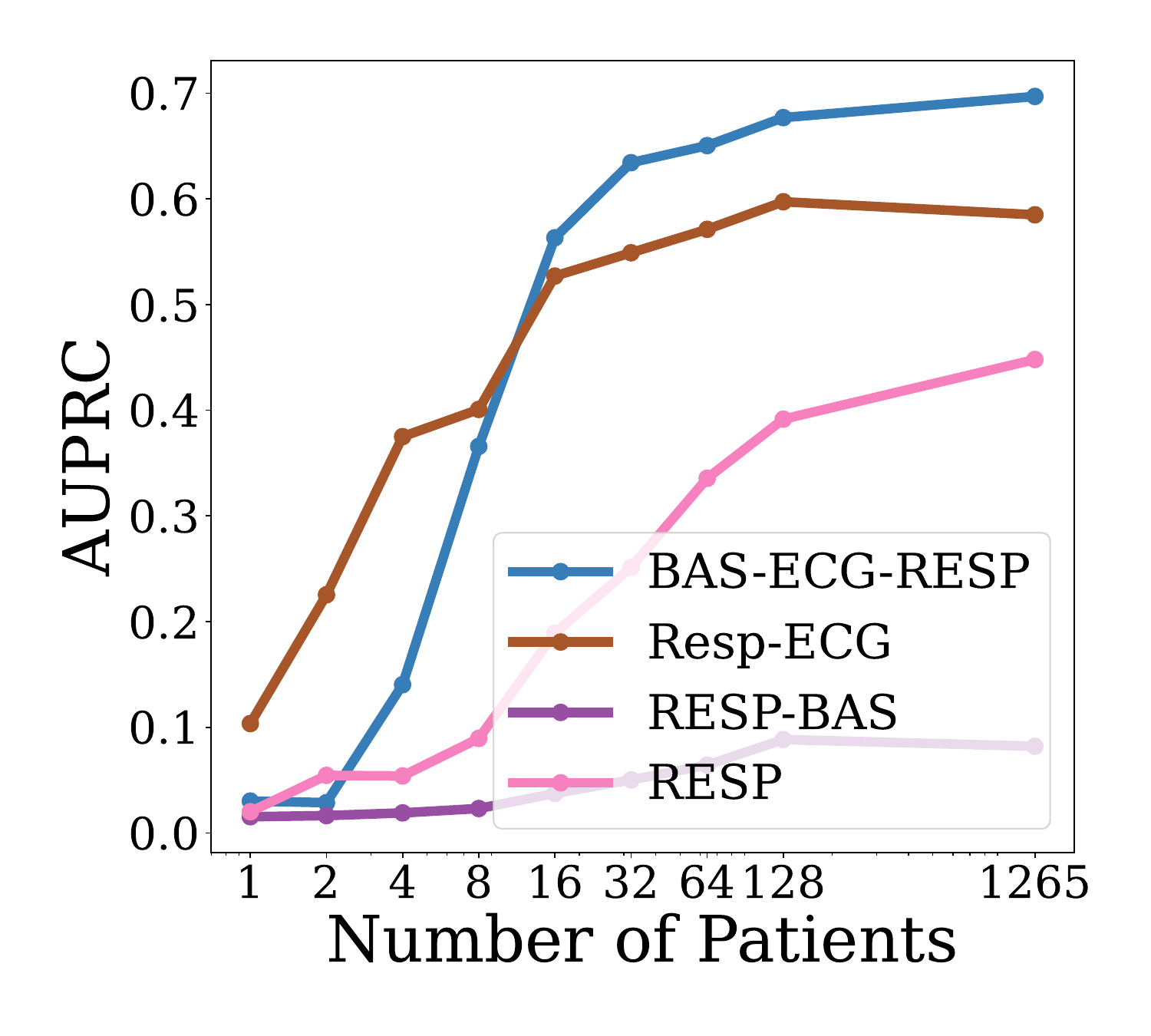}
        \caption{SDB AUPRC}
    \end{subfigure}
    \caption{Ablation few shot plot. The x-axis represents number of patients that the model was trained on and y-axis represents performance metrics AUROC and AUPRC. We select embedding from $k$ number of patients to train a logistic regression model. The last shot (1265) is the total size of our training dataset. The other models (Resp-ECG, Resp-BAS, BAS-ECG) represents the model pretrained using only 2 modalities. Finally, BAS and RESP represents models pretrained with only 1 modality. For each shot, we average the performance across 3 replicates.}
    \label{fig:ablation_modalities}
\end{figure*}

Finally, we conducted ablation studies to analyze how the number and type of modalities used during pretraining impacts downstream task performance. We pretrained models using 3 modalities (BAS-ECG-Respiratory signals), 2 modalities (BAS-Respiratory and BAS-ECG, and ECG-Respiratory), and individual modalities (BAS and Respiratory) with CL. The 3-modality model used leave-one-out CL, which was our best performing model. The 2-modality models used a similar contrastive approach, pairing clips from different modalities in the same 30-second window as positives. For single modalities, adjacent 30-second clips were treated as positive pairs.

For evaluation, we extracted BAS embeddings from all pretrained models and trained logistic regression classifiers for sleep stage classification, a common application of BAS signals. Similarly, for SDB detection, we extracted respiratory embeddings and trained logistic regression models, as respiratory data is typically used for this task. This enables a fair comparison to evaluate which pretraining strategy produces the most useful embeddings for these modalities and applications.

The result of our experiment is shown in \cref{fig:ablation_modalities}. We found pretraining with 3 modalities clearly helped performance across both sleep stage and SDB scoring tasks, with the 3-modality model achieving higher AUPRC for SDB detection in particular. The BAS-ECG and Respiratory-ECG models also performed well, suggesting ECG helps enrich representations of other signals during pretraining. In contrast, single modality models consistently underperformed. Interestingly, certain paired modalities like BAS-Respiratory did not improve performance as much as models incorporating ECG. This indicates the modalities paired during pretraining significantly impact downstream utility of embeddings. Further analysis of how different modality combinations impact representation learning merits exploration in future work.

\begin{table*}[htbp]
    \centering
    \caption{External validation of {\model} on Physionet Computing in Cardiology 2018 challenge \cite{ghassemi2018you}. {\model} was only pretrained on our internal sleep data. Supervised CNN was trained end to end on the external database to classify sleep stages. Test size: 100 participants. $\pm$ represents 95\% confidence intervals.}
    \label{tab:external_validation}
    \begin{tabular}{@{}lccc|ccc@{}}
        \toprule
        & \multicolumn{3}{c|}{\textbf{SleepFM}} & \multicolumn{3}{c}{\textbf{Supervised CNN}} \\
        \cmidrule(r){2-4} \cmidrule(l){5-7}
        & \textbf{AUROC} & \textbf{AUPRC} & \textbf{F1} & \textbf{AUROC} & \textbf{AUPRC} & \textbf{F1}\\
        \midrule
        Wake & $0.966_{\pm .001}$ & $0.867_{\pm .003}$ & $0.790_{\pm .001}$ & $0.867_{\pm .002}$ & $0.614_{\pm .004}$ & $ 0.514_{\pm .002}$ \\
        Stage 1 & $0.830_{\pm .004}$ & $0.471_{\pm .002}$ & $0.439_{\pm .002}$ & $0.709_{\pm .003}$ & $0.305_{\pm .002}$ & $0.006_{\pm .000}$ \\
        Stage 2 & $0.902_{\pm .002}$ & $0.857_{\pm .004}$ & $0.793_{\pm .001}$ & $0.843_{\pm .001}$ & $0.784_{\pm .004}$ & $0.694_{\pm .001}$ \\
        Stage 3 & $0.971_{\pm .001}$ & $0.821_{\pm .004}$ & $0.743_{\pm .001}$ & $0.925_{\pm .002}$ & $0.471_{\pm .002}$ & $0.244_{\pm .001}$ \\
        REM & $0.950_{\pm .002}$ & $0.778_{\pm .005}$ & $0.717_{\pm .002}$ & $0.872_{\pm .003}$ & $0.592_{\pm .002}$ & $0.355_{\pm .001}$ \\
        \midrule
        \textbf{Macro Avg} & $0.924$ & $0.759$ & $0.700$ & $0.843$ & $0.553$ & $0.363$ \\
        \bottomrule
    \end{tabular}
\end{table*}

\subsection{External Validation}
\label{subsection_external_validation}

To evaluate the performance of our model, {\model}, on data from an external site not seen during the pretraining stage, we utilized the publicly available dataset from the Physionet Computing in Cardiology 2018 Challenge \cite{ghassemi2018you}. {\model} was pretrained exclusively on our internal sleep data. For comparison, we also trained a supervised CNN end-to-end on the external dataset to classify sleep stages.

Table \ref{tab:external_validation} presents the results of this external validation. The test set comprised 100 participants, and the metrics reported include AUROC, AUPRC, and F1 score, each accompanied by 95\% confidence intervals.

The {\model} demonstrated superior performance across all sleep stages compared to the supervised CNN. Specifically, {\model} achieved an overall macro-average AUROC of 0.924, AUPRC of 0.759, and F1 score of 0.700. In contrast, the supervised CNN's macro-average metrics were lower, with an AUROC of 0.843, AUPRC of 0.553, and F1 score of 0.363.

The primary takeaway from these results is that {\model} generalizes well to external sites, despite not being exposed to the dataset during the pretraining phase. Additionally, the configuration of EEG channels differs between our site and the CinC dataset. Despite these differences, our model demonstrated robust generalization and adaptation to the new site, showcasing its potential for broader applicability beyond the conditions for which it was specifically trained. This highlights the strength of our approach in handling variations in data acquisition protocols across different sites, a crucial factor for the real-world deployment of sleep analysis models.

%% file: 5_Discussion_Conclusion.tex
\section{Discussion and Conclusion}
\label{section_discussion}

Our study leverages multi-modal PSG data and representation learning techniques to enhance the identification of sleep events, contributing significantly to the field of sleep medicine. The primary contributions include the development and evaluation of a multi-modal contrastive learning (CL) model on a dataset comprising 14,000 participants and over 100,000 hours of sleep data.

Our model demonstrated strong performance across various tasks, including demographic attributes classification, retrieval analysis, sleep stage classification, and sleep-disordered breathing (SDB) event detection, outperforming end-to-end trained CNNs. The methodology centers on two CL approaches: leave-one-out and pairwise. Both approaches effectively unified BAS, ECG, and respiratory signal representations, proving effective in limited data scenarios. Notably, we found that pairwise CL is better suited for cross-modality retrieval, while leave-one-out CL excels in learning representations for downstream sleep stage and SDB classification. This superiority might be attributed to leave-one-modality-out training, which encourages the model to learn a more integrated representation of different modalities.

Moreover, the external validation of {\model} highlights the potential of our approach to be broadly applied in diverse clinical settings, enhancing its utility and impact in sleep research and medicine. This underscores the robustness and versatility of our model, suggesting its capability to handle variations in data acquisition protocols across different sites, a crucial factor for real-world deployment.

\textbf{Future Work.} Despite its achievements, our study has limitations. We primarily trained and evaluated on one institution's sleep data; extensively evaluating the model's generalizability to other institutions is an important direction of future work. We showed that our model works well across different gender and age groups, which is a promising sign of its robustness.
Additionally, while we focused on sleep stage and SDB detection, exploring other tasks like arousal detection, periodic leg movements, and diseases such as narcolepsy could provide a more comprehensive clinical assessment. Moreover, it will be interesting to try our multiple other self-supervised learning (SSL) methods, to see which method actually performs best for this task. Our goal for future work include: (1) pretraining a multi-site, multi-modal foundation model for sleep using diverse PSG data, (2) careful selection and weighting of modalities and handling missing channels, (3) expanding evaluation to more clinically meaningful tasks beyond sleep stage and SDB, and (4) experimenting with multiple other SSL methods.

%% file: 6_Appendix.tex
\section{Appendix}
\label{section_appendix}

\subsection{Data Description}
\label{subsection_data_description_supplementary}

\begin{figure}[htbp]
    \centering
    \includegraphics[width=0.90\textwidth]{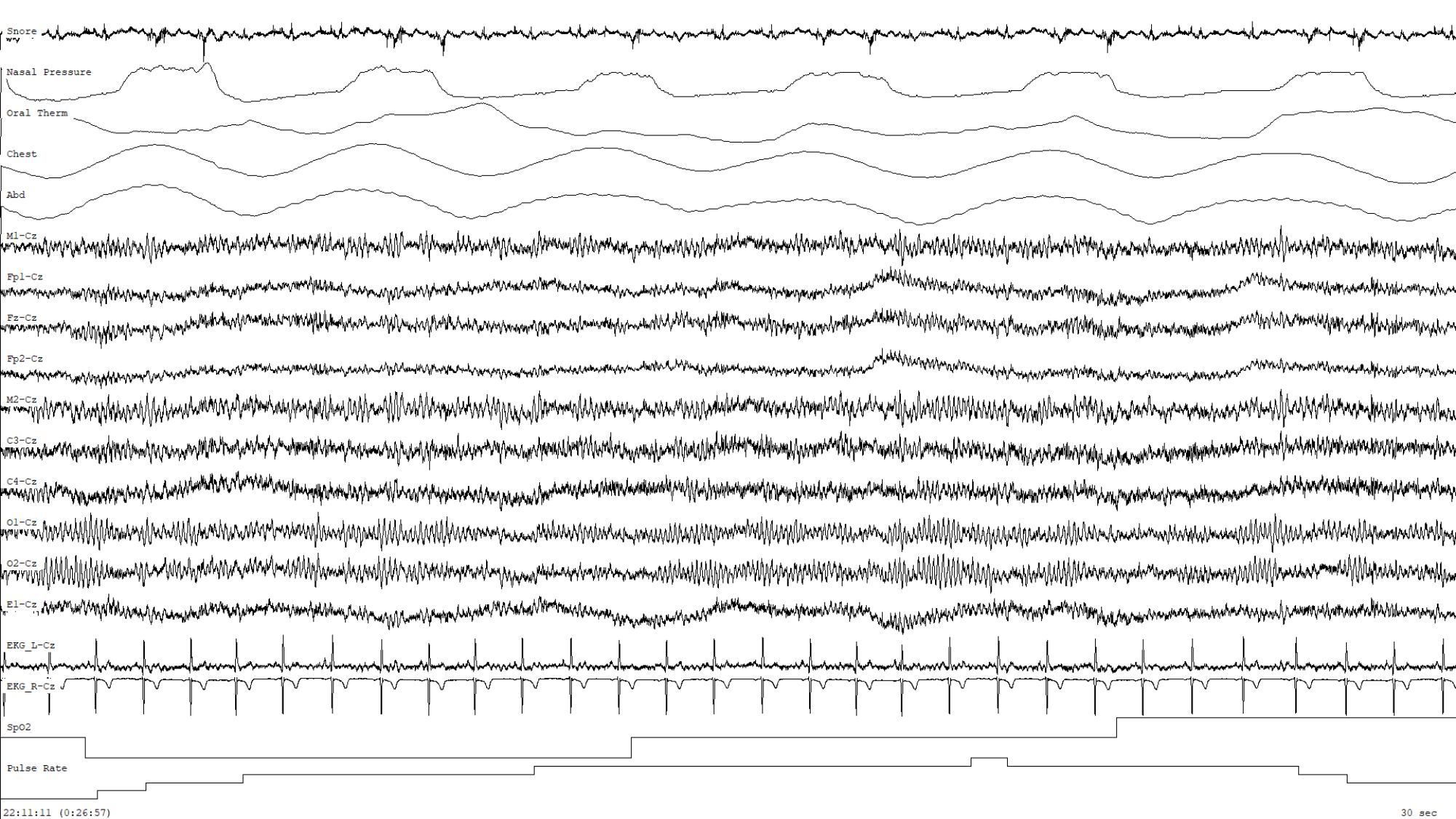}
    \caption{30-second clip of raw patient data. The x-axis is time and y-axis is different channels across all three modalities: BAS, ECG, and Respiratory.}
    \label{fig:raw_data}
\end{figure}

\begin{figure}[htbp]
    \centering
    \includegraphics[width=0.80\textwidth]{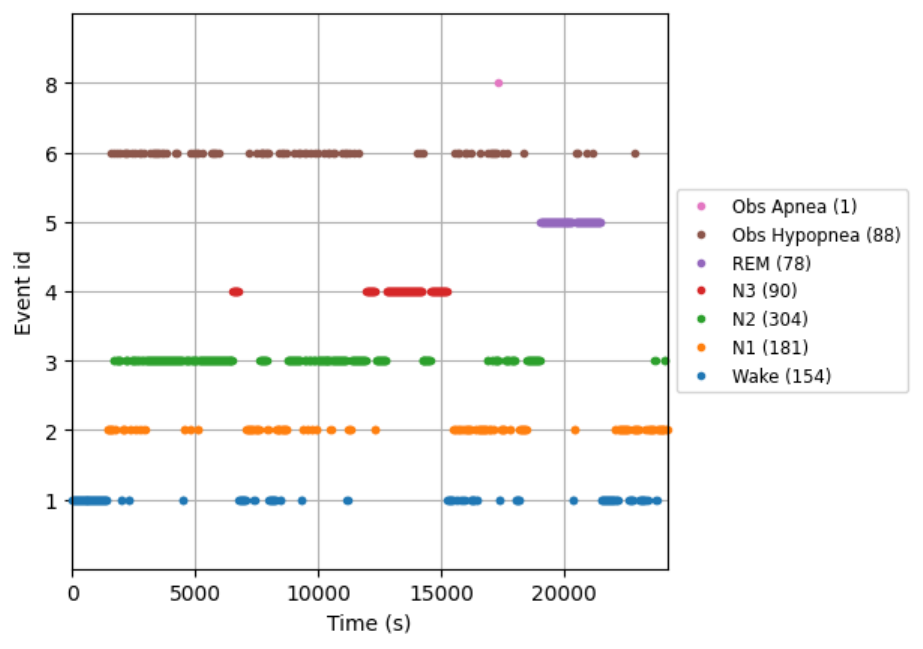}
    \caption{Distribution of events across an entire patient sleep. The x-axis represents approximately 8 hours in seconds, and y-axis is distribution of different sleep events during the entire duration of sleep. N1, N2, N3 refers to Sleep Stage 1, 2, and 3 respectively. Obs Hypopnea and Obs SDB are types of SDBs.}
    \label{fig:events}
\end{figure}

In Figure \ref{fig:raw_data}, we see a 30 second clip of our raw data for all 19 channels across 3 modalities. Figure \ref{fig:events} shows the distribution of various events across the entire sleep duration for a participant. To ensure the protection of participants' Protected Health Information (PHI), all data has been de-identified. 

\subsection{Embedding Model}
\label{subsection_embedding_model_supplementary}

Our EfficientNet model architecture is divided into multiple stages. The first stage (Stage 1) consists of a \texttt{Conv1d} layer with an input channel size of \texttt{in\_channel} and an output channel size of 32. This layer uses a $3 \times 1$ kernel, a stride of 2, padding of 1, and dilation of 1.

After the \texttt{Conv1d} layer, we have a batch normalization layer with an output channel size of 32 as well. Following this, stages 2 to 8 consist of MobileNet blocks, each of which stack together multiple Bottleneck modules. The output channel sizes for each stage are specified by the \texttt{channels} parameter, with the default setting being \texttt{[32, 16, 24, 40, 80, 112, 192, 320, 1280]}. However, these channel sizes are reduced compared to the original EfficientNet to improve runtime efficiency and minimize complexity for the time-series data processing task.

The depth of the model, i.e., the number of Bottleneck modules in each MobileNet block (i.e. layers) in each stage, is controlled by the \texttt{depth} parameter. The number of layers in each stage are \texttt{[1, 2, 2, 3, 3, 3, 3]}.

The model includes two pooling layers with a kernel size of 3, a stride of 1, and padding of 1. The first max-pooling layer is applied after Stage 3, and the second adaptive average pooling layer is applied after Stage 9. A dropout layer with a rate of 0.5 is used before the final fully connected output layer and ReLU activation for regularization. Dropout layers are also used in each of the Bottleneck modules.

The expansion factor for the bottleneck blocks within the MBConv modules is set to 6, as per the MobileNetV2 architecture \cite{sandler2018mobilenetv2}. For a detailed understanding of the architecture of the Bottleneck and EfficientNet model, we refer readers to the original papers \cite{sandler2018mobilenetv2, tan2019efficientnet}.

\subsection{Training Details}
\label{subsection_training_details_supplementary}

All model training was executed on a single NVIDIA Tesla V100S GPU with 32GB of memory. Each pretraining epoch consumed approximately 4 hours, while baseline supervised training required roughly 2 hours on the same GPU. Table \ref{tab:hyperparameters_pretrained} and \ref{tab:hyperparameters_lr} lists the hyperparameters we used in our training runs. 

\begin{table}[htbp]
  \centering
  \caption{Hyperparameters for Pretraining and end-to-end CNN training}
  \label{tab:hyperparameters_pretrained}
  \begin{tabular}{lccc}
    \toprule
    \textbf{Hyperparameter} & \textbf{Value} \\
    \midrule
    Learning Rate & 0.01 \\
    Batch Size & 32 \\
    lr step period & 5 \\
    epochs & 20 \\
    momentum & 0.9 \\
    Temperature (init) & 0.0 \\
    Dropout & 0.5 \\
    \bottomrule
  \end{tabular}
\end{table}

\begin{table}[htbp]
  \centering
  \caption{Hyperparameters for logistic regression training during downstream classifications.}
  \label{tab:hyperparameters_lr}
  \begin{tabular}{lccc}
    \toprule
    \textbf{Hyperparameter} & \textbf{Value} \\
    \midrule
    penalty & L2 \\
    max iter & 10000 \\
    class weight & balanced \\
    solver & lbfgs \\
    \bottomrule
  \end{tabular}
\end{table}

\subsection{Additional Results}
\label{subsection_additional_results}

\begin{table*}[htbp]
    \centering
    \caption{Sleep stage classification metrics for model trained with leave-one-out CL. After having trained the model with all three modalities, we extract embeddings for each modality separately and train a logistic regression with each modality to identify sleep stages. $\pm$ represents 95\% confidence intervals.}
    \label{tab:modality_sleep_stage_classification_metrics_leave_one_out}
    \begin{tabular}{@{}lccc|ccc@{}}
        \toprule
        & \multicolumn{3}{c|}{\textbf{AUROC}} & \multicolumn{3}{c}{\textbf{AUPRC}} \\
        \cmidrule(r){2-4} \cmidrule(l){5-7}
        & \textbf{ECG} & \textbf{Respiratory} & \textbf{BAS} & \textbf{ECG} & \textbf{Respiratory} & \textbf{BAS}\\
        \midrule
        Wake & $0.934_{\pm .001}$ & $0.846_{\pm .001}$ & $0.942_{\pm .001}$ & $0.829_{\pm .004}$ & $0.652_{\pm .003}$ & $0.857_{\pm .002}$ \\
        Stage 1 & $0.786_{\pm .002}$ & $0.676_{\pm .002}$ & $0.801_{\pm .002}$ & $0.193_{\pm .002}$ & $0.127_{\pm .001}$ & $0.211_{\pm .003}$ \\
        Stage 2 & $0.874_{\pm .001}$ & $0.728_{\pm .001}$ & $0.888_{\pm .001}$ & $0.860_{\pm .001}$ & $0.708_{\pm .001}$ & $0.873_{\pm .001}$ \\
        Stage 3 & $0.919_{\pm .001}$ & $0.788_{\pm .001}$ & $0.927_{\pm .001}$ & $0.638_{\pm .003}$ & $0.307_{\pm .002}$ & $0.679_{\pm .002}$ \\
        REM & $0.939_{\pm .001}$ & $0.789_{\pm .001}$ & $0.944_{\pm .001}$ & $0.745_{\pm .003}$ & $0.388_{\pm .003}$ & $0.724_{\pm .003}$ \\
        \midrule
        \textbf{Macro Avg} & $0.891$ & $0.765$ & $0.900$ & $0.436$ & $0.484$ & $0.669$ \\
        \bottomrule
    \end{tabular}
\end{table*}

\begin{table*}[htbp]
    \centering
    \caption{SDB classification metrics for model trained with leave-one-out CL. After having trained the model with all three modalities, we extract embeddings for each modality separately and train a logistic regression with each modality to identify SDB. $\pm$ represents 95\% confidence intervals.}
    \label{tab:modality_SDB_classification_metrics_leave_one_out}
    \begin{tabular}{@{}lccc@{}}
        \toprule
        & \textbf{ECG} & \textbf{Respiratory} & \textbf{BAS} \\
        \midrule
        AUROC & $0.735_{\pm .004}$ & $0.925_{\pm .002}$ & $0.735_{\pm .004}$ \\
        AUPRC & $0.040_{\pm .001}$ & $0.697_{\pm .006}$ & $0.040_{\pm .001}$ \\
        \bottomrule
    \end{tabular}
    
\end{table*}

\begin{table*}[htbp]
    \centering
    \caption{Sleep stage classification metrics for model trained with pairwise CL. After having trained the model with all three modalities, we extract embeddings for each modality separately and train a logistic regression with each modality to identify sleep stages. $\pm$ represents 95\% confidence intervals.}
    \label{tab:modality_sleep_stage_classification_metrics_pairwise}
    \begin{tabular}{@{}lccc|ccc@{}}
        \toprule
        & \multicolumn{3}{c|}{\textbf{AUROC}} & \multicolumn{3}{c}{\textbf{AUPRC}} \\
        \cmidrule(r){2-4} \cmidrule(l){5-7}
        & \textbf{ECG} & \textbf{Respiratory} & \textbf{BAS} & \textbf{ECG} & \textbf{Respiratory} & \textbf{BAS}\\
        \midrule
        Wake & $0.917_{\pm .001}$ & $0.821_{\pm .001}$ & $0.925_{\pm .001}$ & $0.782_{\pm .002}$ & $0.621_{\pm .002}$ & $0.816_{\pm .001}$ \\
        Stage 1 & $0.766_{\pm .002}$ & $0.661_{\pm .002}$ & $0.772_{\pm .002}$ & $0.167_{\pm .002}$ & $0.116_{\pm .001}$ & $0.174_{\pm .002}$ \\
        Stage 2 & $0.848_{\pm .001}$ & $0.695_{\pm .001}$ & $0.857_{\pm .001}$ & $0.841_{\pm .001}$ & $0.675_{\pm .001}$ & $0.845_{\pm .001}$ \\
        Stage 3 & $0.911_{\pm .001}$ & $0.777_{\pm .001}$ & $0.917_{\pm .001}$ & $0.601_{\pm .002}$ & $0.296_{\pm .003}$ & $0.614_{\pm .003}$ \\
        REM & $0.872_{\pm .001}$ & $0.649_{\pm .001}$ & $0.880_{\pm .001}$ & $0.526_{\pm .003}$ & $0.200_{\pm .003}$ & $0.522_{\pm .002}$ \\
        \midrule
        \textbf{Macro Avg} & $0.862$ & $0.720$ & $0.870$ & $0.583$ & $0.381$ & $0.594$ \\
        \bottomrule
    \end{tabular}
\end{table*}

\begin{table*}[htbp]
    \centering
    \caption{SDB classification metrics for model trained with pairwise CL. After having trained the model with all three modalities, we extract embeddings for each modality separately and train a logistic regression with each modality to identify SDB. $\pm$ represents 95\% confidence intervals.}
    \label{tab:modality_SDB_classification_metrics_pariwise}
    \begin{tabular}{@{}lccc@{}}
        \toprule
        & \textbf{ECG} & \textbf{Respiratory} & \textbf{BAS} \\
        \midrule
        AUROC & $0.698_{\pm .003}$ & $0.893_{\pm .003}$ & $0.706_{\pm .004}$ \\
        AUPRC & $0.029_{\pm .001}$ & $0.601_{\pm .006}$ & $0.030_{\pm .001}$ \\
        \bottomrule
    \end{tabular}
    
\end{table*}

\begin{table*}[htbp]
    \centering
    \caption{Age classification metrics for model trained with leave-one-out CL. After having trained the model with all three modalities, we extract embeddings for each modality separately and train a logistic regression with each modality to identify age groups. $\pm$ represents 95\% confidence intervals.}
    \label{tab:modality_age_classification_metrics_leave_one_out}
    \begin{tabular}{@{}lccc|ccc@{}}
        \toprule
        & \multicolumn{3}{c|}{\textbf{AUROC}} & \multicolumn{3}{c}{\textbf{AUPRC}} \\
        \cmidrule(r){2-4} \cmidrule(l){5-7}
        & \textbf{ECG} & \textbf{Respiratory} & \textbf{BAS} & \textbf{ECG} & \textbf{Respiratory} & \textbf{BAS}\\
        \midrule
        0-18 & $0.977_{\pm .001}$ & $0.965_{\pm .001}$ & $0.969_{\pm .001}$ & $0.921_{\pm .001}$ & $0.883_{\pm .003}$ & $0.911_{\pm .001}$ \\
        18-35 & $0.833_{\pm .001}$ & $0.789_{\pm .001}$ & $0.755_{\pm .002}$ & $0.493_{\pm .003}$ & $0.455_{\pm .003}$ & $0.380_{\pm .003}$ \\
        35-50 & $0.774_{\pm .001}$ & $0.722_{\pm .001}$ & $0.686_{\pm .001}$ & $0.516_{\pm .002}$ & $0.458_{\pm .003}$ & $0.424_{\pm .002}$ \\
        50+ & $0.905_{\pm .001}$ & $0.873_{\pm .001}$ & $0.813_{\pm .001}$ & $0.843_{\pm .001}$ & $0.780_{\pm .001}$ & $0.685_{\pm .002}$ \\
        \midrule
        \textbf{Macro Avg} & $0.872$ & $0.837$ & $0.805$ & $0.693$ & $0.644$ & $0.600$ \\
        \bottomrule
    \end{tabular}
\end{table*}

\begin{table*}[htbp]
    \centering
    \caption{Gender classification metrics for model trained with leave-one-out CL. After having trained the model with all three modalities, we extract embeddings for each modality separately and train a logistic regression with each modality to identify gender. $\pm$ represents 95\% confidence intervals.}
    \label{tab:modality_gender_classification_metrics_leave_one_out}
    \begin{tabular}{@{}lccc@{}}
        \toprule
        & \textbf{ECG} & \textbf{Respiratory} & \textbf{BAS} \\
        \midrule
        AUROC & $0.829_{\pm .001}$ & $0.790_{\pm .002}$ & $0.778_{\pm .001}$ \\
        AUPRC & $0.754_{\pm .001}$ & $0.710_{\pm .003}$ & $0.713_{\pm .002}$ \\
        \bottomrule
    \end{tabular}
    
\end{table*}

\begin{table*}[htbp]
    \centering
    \caption{Age classification metrics for model trained with pairwise CL. After having trained the model with all three modalities, we extract embeddings for each modality separately and train a logistic regression with each modality to identify age groups. $\pm$ represents 95\% confidence intervals.}
    \label{tab:modality_age_classification_metrics_pairwise}
    \begin{tabular}{@{}lccc|ccc@{}}
        \toprule
        & \multicolumn{3}{c|}{\textbf{AUROC}} & \multicolumn{3}{c}{\textbf{AUPRC}} \\
        \cmidrule(r){2-4} \cmidrule(l){5-7}
        & \textbf{ECG} & \textbf{Respiratory} & \textbf{BAS} & \textbf{ECG} & \textbf{Respiratory} & \textbf{BAS}\\
        \midrule
        0-18 & $0.969_{\pm .001}$ & $0.962_{\pm .001}$ & $0.963_{\pm .001}$ & $0.908_{\pm .001}$ & $0.883_{\pm .001}$ & $0.897_{\pm .001}$ \\
        18-35 & $0.786_{\pm .001}$ & $0.769_{\pm .001}$ & $0.767_{\pm .001}$ & $0.422_{\pm .002}$ & $0.455_{\pm .003}$ & $0.389_{\pm .002}$ \\
        35-50 & $0.712_{\pm .002}$ & $0.702_{\pm .001}$ & $0.706_{\pm .002}$ & $0.441_{\pm .002}$ & $0.458_{\pm .003}$ & $0.436_{\pm .002}$ \\
        50+ & $0.865_{\pm .001}$ & $0.841_{\pm .001}$ & $0.840_{\pm .001}$ & $0.722_{\pm .002}$ & $0.780_{\pm .001}$ & $0.742_{\pm .001}$ \\
        \midrule
        \textbf{Macro Avg} & $0.832$ & $0.818$ & $0.818$ & $0.634$ & $0.617$ & $0.615$ \\
        \bottomrule
    \end{tabular}
\end{table*}

\begin{table*}[htbp]
    \centering
    \caption{Gender classification metrics for model trained with pairwise CL. After having trained the model with all three modalities, we extract embeddings for each modality separately and train a logistic regression with each modality to identify gender. $\pm$ represents 95\% confidence intervals.}
    \label{tab:modality_gender_classification_metrics_pairwise}
    \begin{tabular}{@{}lccc@{}}
        \toprule
        & \textbf{ECG} & \textbf{Respiratory} & \textbf{BAS} \\
        \midrule
        AUROC & $0.795_{\pm .001}$ & $0.746_{\pm .001}$ & $0.765_{\pm .001}$ \\
        AUPRC & $0.722_{\pm .001}$ & $0.676_{\pm .002}$ & $0.702_{\pm .002}$ \\
        \bottomrule
    \end{tabular}
    
\end{table*}

\begin{table*}[t]
    \centering
    \caption{Sleep Stage Classification stratified by age group.}
    \label{tab:sleep_stage_classification_age_macro}
    \begin{tabular}{@{}lcc|cc@{}}
        \toprule
        & \multicolumn{2}{c|}{\textbf{Macro AUROC}} & \multicolumn{2}{c}{\textbf{Macro AUPRC}} \\
        \cmidrule(r){2-3} \cmidrule(l){3-5}
        & \textbf{Leave-One-Out} & \textbf{Pairwise} & \textbf{Leave-One-Out} & \textbf{Pairwise} \\
        \midrule
        0-18 & $0.890$  & $0.849$ & $0.665$ & $0.594$ \\
        18-35 & $0.911$ & $0.883$ & $0.702$ & $0.624$ \\
        35-50 & $0.897$  & $0.867$ & $0.630$ & $0.559$ \\
        50+ & $0.895$  & $0.861$ & $0.616$ & $0.530$ \\
        \bottomrule
    \end{tabular}
\end{table*}

\begin{table*}[t]
    \centering
    \caption{Sleep Stage Classification stratified by gender.}
    \label{tab:sleep_stage_classification_gender_macro}
    \begin{tabular}{@{}lcc|cc@{}}
        \toprule
        & \multicolumn{2}{c|}{\textbf{Macro AUROC}} & \multicolumn{2}{c}{\textbf{Macro AUPRC}} \\
        \cmidrule(r){2-3} \cmidrule(l){3-5}
        & \textbf{Leave-One-Out} & \textbf{Pairwise} & \textbf{Leave-One-Out} & \textbf{Pairwise} \\
        \midrule
        Male & $0.899$  & $0.869$ & $0.674$ & $0.594$ \\
        Female & $0.910$ & $0.880$ & $0.693$ & $0.621$ \\
        \bottomrule
    \end{tabular}
\end{table*}

\begin{table*}[t]
    \centering
    \caption{SDB classification metrics stratified by age group.}
    \label{tab:SDB_classification_age}
    \begin{tabular}{@{}lcc|cc@{}}
        \toprule
        & \multicolumn{2}{c|}{\textbf{AUROC}} & \multicolumn{2}{c}{\textbf{AUPRC}} \\
        \cmidrule(r){2-3} \cmidrule(l){3-5}
        & \textbf{Leave-One-Out} & \textbf{Pairwise} & \textbf{Leave-One-Out} & \textbf{Pairwise} \\
        \midrule
        0-18 & $0.93_{\pm 0.01}$ & $0.86_{\pm 0.03}$ & $0.56_{\pm 0.04}$ & $0.35_{\pm 0.04}$ \\
        18-35 & $0.94_{\pm 0.01}$ & $0.90_{\pm 0.01}$ & $0.69_{\pm 0.02}$ & $0.61_{\pm 0.03}$ \\
        35-50 & $0.94_{\pm 0.01}$ & $0.89_{\pm 0.01}$ & $0.73_{\pm 0.01}$ & $0.63_{\pm 0.02}$ \\
        50+ & $0.94_{\pm 0.01}$ & $0.90_{\pm 0.01}$ & $0.73_{\pm 0.01}$ & $0.60_{\pm 0.01}$ \\
        \bottomrule
    \end{tabular}
\end{table*}

\begin{table*}[t]
    \centering
    \caption{SDB classification metrics stratified by gender.}
    \label{tab:SDB_classification_gender}
    \begin{tabular}{@{}lcc|cc@{}}
        \toprule
        & \multicolumn{2}{c|}{\textbf{AUROC}} & \multicolumn{2}{c}{\textbf{AUPRC}} \\
        \cmidrule(r){2-3} \cmidrule(l){3-5}
        & \textbf{Leave-One-Out} & \textbf{Pairwise} & \textbf{Leave-One-Out} & \textbf{Pairwise} \\
        \midrule
        Male & $0.94_{\pm 0.01}$ & $0.90_{\pm 0.01}$ & $0.73_{\pm 0.01}$ & $0.61_{\pm 0.01}$ \\
        Female & $0.95_{\pm 0.01}$ & $0.91_{\pm 0.01}$ & $0.70_{\pm 0.01}$ & $0.59_{\pm 0.01}$ \\
        \bottomrule
    \end{tabular}
\end{table*}

\begin{table}[t]
    \centering
    \caption{Sleep stage classification AUROC metrics for model trained with leave-one-out CL, stratified by different age groups.}
    \label{tab:sleep_stage_classification_auroc_leave_one_out_age_split}
    \begin{tabular}{@{}lcccc@{}}
        \toprule
        & \textbf{0-18} & \textbf{18-35} & \textbf{35-50} & \textbf{50+} \\
        \midrule
        Wake & $0.937_{\pm 0.002}$ & $0.939_{\pm 0.001}$ & $0.938_{\pm 0.001}$ & $0.944_{\pm 0.001}$ \\
        Stage 1 & $0.805_{\pm 0.006}$ & $0.831_{\pm 0.003}$ & $0.808_{\pm 0.003}$ & $0.793_{\pm 0.002}$ \\
        Stage 2 & $0.861_{\pm 0.002}$ & $0.900_{\pm 0.001}$ & $0.888_{\pm 0.002}$ & $0.889_{\pm 0.001}$ \\
        Stage 3 & $0.906_{\pm 0.001}$ & $0.932_{\pm 0.002}$ & $0.902_{\pm 0.002}$ & $0.902_{\pm 0.002}$ \\
        REM & $0.941_{\pm 0.002}$ & $0.956_{\pm 0.001}$ & $0.950_{\pm 0.001}$ & $0.949_{\pm 0.001}$ \\
        \midrule
        \textbf{Avg} & $0.890$ & $0.911$ & $0.897$ & $0.895$ \\
        \bottomrule
    \end{tabular}
\end{table}

\begin{table}[t]
    \centering
    \caption{Sleep stage classification AUPRC metrics for model trained with leave-one-out CL, stratified by different age groups.}
    \label{tab:sleep_stage_classification_auprc_leave_one_out_age_split}
    \begin{tabular}{@{}lcccc@{}}
        \toprule
        & \textbf{0-18} & \textbf{18-35} & \textbf{35-50} & \textbf{50+} \\
        \midrule
        Wake & $0.809_{\pm 0.005}$ & $0.859_{\pm 0.004}$ & $0.843_{\pm 0.003}$ & $0.872_{\pm 0.002}$ \\
        Stage 1 & $0.163_{\pm 0.008}$ & $0.290_{\pm 0.006}$ & $0.236_{\pm 0.005}$ & $0.235_{\pm 0.004}$ \\
        Stage 2 & $0.812_{\pm 0.003}$ & $0.890_{\pm 0.002}$ & $0.879_{\pm 0.001}$ & $0.863_{\pm 0.002}$ \\
        Stage 3 & $0.818_{\pm 0.004}$ & $0.696_{\pm 0.004}$ & $0.406_{\pm 0.007}$ & $0.325_{\pm 0.005}$ \\
        REM & $0.725_{\pm 0.007}$ & $0.775_{\pm 0.006}$ & $0.787_{\pm 0.004}$ & $0.786_{\pm 0.004}$ \\
        \midrule
        \textbf{Avg} & $0.665$ & $0.702$ & $0.630$ & $0.616$ \\
        \bottomrule
    \end{tabular}
\end{table}

\begin{table*}[t]
    \centering
    \caption{Sleep stage classification metrics for model trained with leave-one-out CL. The performance is stratified by different gender groups.}
    \label{tab:sleep_stage_classification_leave_one_out_gender_split}
    \begin{tabular}{@{}lcc|cc@{}}
        \toprule
        & \multicolumn{2}{c|}{\textbf{AUROC}} & \multicolumn{2}{c}{\textbf{AUPRC}} \\
        \cmidrule(r){2-3} \cmidrule(l){3-5}
        & \textbf{Male} & \textbf{Female} & \textbf{Male} & \textbf{Female} \\
        \midrule
        Wake & $0.937_{\pm 0.001}$  & $0.949_{\pm 0.001}$ & $0.844_{\pm 0.002}$ & $0.872_{\pm 0.002}$ \\
        Stage 1 & $0.805_{\pm 0.002}$ & $0.824_{\pm 0.002}$ & $0.251_{\pm 0.004}$ & $0.225_{\pm 0.004}$ \\
        Stage 2 & $0.887_{\pm 0.001}$  & $0.890_{\pm 0.001}$ & $0.867_{\pm 0.001}$ & $0.870_{\pm 0.001}$ \\
        Stage 3 & $0.919_{\pm 0.001}$  & $0.934_{\pm 0.001}$ & $0.635_{\pm 0.005}$ & $0.729_{\pm 0.004}$ \\
        REM & $0.944_{\pm 0.001}$  & $0.955_{\pm 0.001}$ & $0.771_{\pm 0.004}$ & $0.767_{\pm 0.002}$ \\
        \midrule
        \textbf{Avg} & $0.899$ & $0.910$ & $0.674$ & $0.693$ \\
        \bottomrule
    \end{tabular}
\end{table*}

\begin{table}[t]
    \centering
    \caption{Sleep stage classification AUROC metrics for model trained with pairwise CL, stratified by different age groups.}
    \label{tab:sleep_stage_classification_auroc_pairwise_age_split}
    \begin{tabular}{@{}lcccc@{}}
        \toprule
        & \textbf{0-18} & \textbf{18-35} & \textbf{35-50} & \textbf{50+} \\
        \midrule
        Wake & $0.919_{\pm 0.002}$ & $0.928_{\pm 0.002}$ & $0.926_{\pm 0.001}$ & $0.926_{\pm 0.001}$ \\
        Stage 1 & $0.712_{\pm 0.009}$ & $0.804_{\pm 0.004}$ & $0.775_{\pm 0.003}$ & $0.758_{\pm 0.003}$ \\
        Stage 2 & $0.827_{\pm 0.002}$ & $0.870_{\pm 0.002}$ & $0.863_{\pm 0.002}$ & $0.861_{\pm 0.002}$ \\
        Stage 3 & $0.891_{\pm 0.002}$ & $0.911_{\pm 0.002}$ & $0.881_{\pm 0.003}$ & $0.891_{\pm 0.002}$ \\
        REM & $0.894_{\pm 0.002}$ & $0.901_{\pm 0.002}$ & $0.891_{\pm 0.002}$ & $0.868_{\pm 0.002}$ \\
        \midrule
        \textbf{Avg} & $0.849$ & $0.883$ & $0.867$ & $0.861$ \\
        \bottomrule
    \end{tabular}
\end{table}

\begin{table}[t]
    \centering
    \caption{Sleep stage classification AUPRC metrics for model trained with pairwise CL, stratified by different age groups.}
    \label{tab:sleep_stage_classification_auprc_pairwise_age_split}
    \begin{tabular}{@{}lcccc@{}}
        \toprule
        & \textbf{0-18} & \textbf{18-35} & \textbf{35-50} & \textbf{50+} \\
        \midrule
        Wake & $0.771_{\pm 0.005}$ & $0.828_{\pm 0.003}$ & $0.813_{\pm 0.003}$ & $0.838_{\pm 0.003}$ \\
        Stage 1 & $0.103_{\pm 0.006}$ & $0.218_{\pm 0.007}$ & $0.191_{\pm 0.004}$ & $0.198_{\pm 0.004}$ \\
        Stage 2 & $0.780_{\pm 0.003}$ & $0.861_{\pm 0.003}$ & $0.857_{\pm 0.002}$ & $0.833_{\pm 0.002}$ \\
        Stage 3 & $0.775_{\pm 0.004}$ & $0.617_{\pm 0.003}$ & $0.340_{\pm 0.009}$ & $0.267_{\pm 0.007}$ \\
        REM & $0.539_{\pm 0.009}$ & $0.597_{\pm 0.006}$ & $0.591_{\pm 0.006}$ & $0.516_{\pm 0.005}$ \\
        \midrule
        \textbf{Avg} & $0.594$ & $0.624$ & $0.559$ & $0.530$ \\
        \bottomrule
    \end{tabular}
\end{table}

\begin{table*}[t]
    \centering
    \caption{Sleep stage classification metrics for model trained with pairwise CL. The performance is stratified by different gender groups.}
    \label{tab:sleep_stage_classification_pairwise_gender_split}
    \begin{tabular}{@{}lcc|cc@{}}
        \toprule
        & \multicolumn{2}{c|}{\textbf{AUROC}} & \multicolumn{2}{c}{\textbf{AUPRC}} \\
        \cmidrule(r){2-3} \cmidrule(l){3-5}
        & \textbf{Male} & \textbf{Female} & \textbf{Male} & \textbf{Female} \\
        \midrule
        Wake & $0.924_{\pm 0.001}$  & $0.932_{\pm 0.001}$ & $0.813_{\pm 0.002}$ & $0.834_{\pm 0.002}$ \\
        Stage 1 & $0.769_{\pm 0.002}$ & $0.791_{\pm 0.002}$ & $0.194_{\pm 0.003}$ & $0.192_{\pm 0.004}$ \\
        Stage 2 & $0.859_{\pm 0.001}$  & $0.861_{\pm 0.001}$ & $0.840_{\pm 0.001}$ & $0.840_{\pm 0.002}$ \\
        Stage 3 & $0.910_{\pm 0.001}$  & $0.922_{\pm 0.001}$ & $0.559_{\pm 0.002}$ & $0.687_{\pm 0.004}$ \\
        REM & $0.882_{\pm 0.001}$  & $0.892_{\pm 0.001}$ & $0.561_{\pm 0.002}$ & $0.554_{\pm 0.005}$ \\
        \midrule
        \textbf{Avg} & $0.869$ & $0.880$ & $0.594$ & $0.621$ \\
        \bottomrule
    \end{tabular}
\end{table*}

\begin{table*}[htbp]
    \centering
    \caption{Sleep stage classification metrics for model trained with supervised CNN individually on each modality. $\pm$ represents 95\% confidence intervals.}
    \label{tab:individual_modality_sleep_stages_supervised_CNN}
    \begin{tabular}{@{}lccc|ccc@{}}
        \toprule
        & \multicolumn{3}{c|}{\textbf{AUROC}} & \multicolumn{3}{c}{\textbf{AUPRC}} \\
        \cmidrule(r){2-4} \cmidrule(l){5-7}
        & \textbf{ECG} & \textbf{Respiratory} & \textbf{BAS} & \textbf{ECG} & \textbf{Respiratory} & \textbf{BAS}\\
        \midrule
        Wake & $0.440_{\pm .004}$ & $0.571_{\pm .001}$ & $0.916_{\pm .001}$ & $0.186_{\pm .001}$ & $0.277_{\pm .002}$ & $0.800_{\pm .002}$ \\
        Stage 1 & $0.478_{\pm .002}$ & $0.564_{\pm .002}$ & $0.736_{\pm .002}$ & $0.063_{\pm .001}$ & $0.087_{\pm .001}$ & $0.146_{\pm .001}$ \\
        Stage 2 & $0.474_{\pm .001}$ & $0.540_{\pm .002}$ & $0.823_{\pm .001}$ & $0.481_{\pm .002}$ & $0.550_{\pm .001}$ & $0.800_{\pm .001}$ \\
        Stage 3 & $0.620_{\pm .001}$ & $0.552_{\pm .002}$ & $0.895_{\pm .001}$ & $0.121_{\pm .001}$ & $0.120_{\pm .002}$ & $0.593_{\pm .004}$ \\
        REM & $0.490_{\pm .002}$ & $0.591_{\pm .002}$ & $0.875_{\pm .001}$ & $0.128_{\pm .003}$ & $0.171_{\pm .002}$ & $0.581_{\pm .003}$ \\
        \midrule
        \textbf{Macro Avg} & $0.500$ & $0.563$ & $0.850$ & $0.195$ & $0.241$ & $0.584$ \\
        \bottomrule
    \end{tabular}
\end{table*}

\begin{table*}[htbp]
    \centering
    \caption{SDB classification metrics for model trained with supervised CNN individually on each modality. $\pm$ represents 95\% confidence intervals.}
    \label{tab:individual_modality_SDB_supervised_CNN}
    \begin{tabular}{@{}lccc@{}}
        \toprule
        & \textbf{ECG} & \textbf{Respiratory} & \textbf{BAS} \\
        \midrule
        AUROC & $0.552_{\pm .004}$ & $0.870_{\pm .003}$ & $0.387_{\pm .004}$ \\
        AUPRC & $0.019_{\pm .001}$ & $0.553_{\pm .003}$ & $0.012_{\pm .001}$ \\
        \bottomrule
    \end{tabular}
    
\end{table*}